\documentclass[10pt,twocolumn,letterpaper]{article}

\usepackage{cvpr}
\usepackage{times}
\usepackage{epsfig}
\usepackage{graphicx}
\usepackage{amsmath}
\usepackage{amssymb}

\usepackage{float}
\usepackage{bbm}
\usepackage{color}
\usepackage{placeins}
\usepackage{color,soul}
\usepackage[utf8]{inputenc}
\usepackage{mathtools}
\usepackage{stackengine}

\usepackage{cvpr}
\usepackage{times}
\usepackage{epsfig}
\usepackage{graphicx}
\usepackage{amsmath}
\usepackage{amssymb}
\usepackage{float}
\usepackage{bbm}
\usepackage{color}
\usepackage{placeins}
\usepackage{color,soul}
\usepackage{tabularx}
\usepackage[shortlabels]{enumitem}

\usepackage{mathtools}

\DeclarePairedDelimiter\floor{\lfloor}{\rfloor}

\newcommand{\mD}{\mathcal{D}}
\newcommand\xrowht[2][0]{\addstackgap[.5\dimexpr#2\relax]{\vphantom{#1}}}

\hypersetup{pagebackref=true,breaklinks=true,letterpaper=true,colorlinks,bookmarks=false}

\cvprfinalcopy 

\def\cvprPaperID{4284} 

\ifcvprfinal\fi

\begin{document}

\title{A Content Transformation Block For Image Style Transfer}

\author{Dmytro Kotovenko \hspace{15pt} 
Artsiom Sanakoyeu \hspace{15pt} 
Pingchuan Ma \hspace{15pt} 
Sabine Lang \hspace{15pt} 
Bj{\"o}rn Ommer \\
Heidelberg Collaboratory for Image Processing,
IWR,
Heidelberg University \\
}

\maketitle
\thispagestyle{empty}

\begin{abstract}
    Style transfer has recently received a lot of attention, since it allows to study fundamental challenges in image understanding and synthesis. Recent work has significantly improved the representation of color and texture and computational speed and image resolution. The explicit transformation of image content has, however, been mostly neglected: while artistic style affects formal characteristics of an image, such as color, shape or texture, it also deforms, adds or removes content details. This paper explicitly focuses on a content-and style-aware stylization of a content image. Therefore, we introduce a content transformation module between the encoder and decoder. Moreover, we utilize similar content appearing in photographs and style samples to learn how style alters content details and we generalize this to other class details. Additionally, this work presents a novel normalization layer critical for high resolution image synthesis. The robustness and speed of our model enables a video stylization in real-time and high definition. We perform extensive qualitative and quantitative evaluations to demonstrate the validity of our approach.

   
\end{abstract}

\section{Introduction}
 
\begin{figure}[t!]
    \centering
    \includegraphics[width=0.4\textwidth,height=0.9\textwidth]{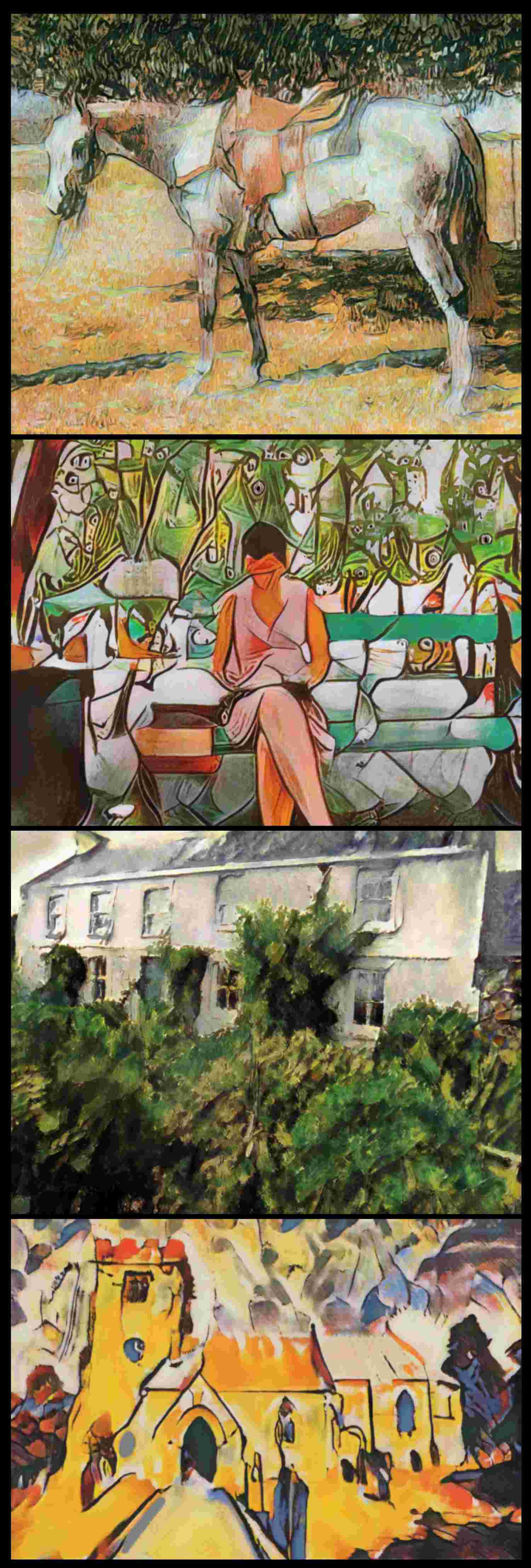}
    \caption{Examples of generated images using our approach in the style of (from top) Vincent van Gogh, Pablo Picasso, Paul Cezanne and Wassily Kandinsky. More stylization examples of images and videos can be found on the \href{https://compvis.github.io/content-targeted-style-transfer/}{project page}\protect\footnotemark .}
    \label{fig:sample_images}
\end{figure}
\footnotetext{\texttt{compvis.github.io/content-targeted-style-transfer/}}

    \begin{figure*}
    \begin{center}
    
        \includegraphics[width=1.0\textwidth,keepaspectratio]{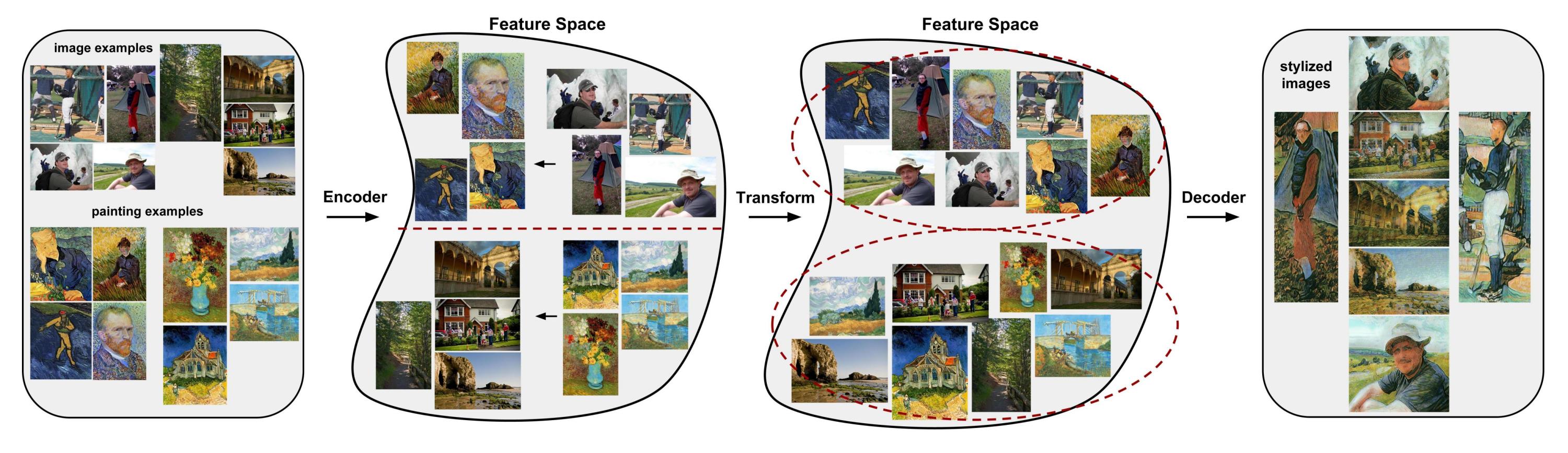}
    \caption{Illustrates the role of the content transformation block. Real input images and painting examples displaying similar content are matched in the feature space. We utilize content similarity to learn how style explicitly alters content details.}
    \label{fig:hr}
    \end{center}   
    \end{figure*}

Style transfer renders the content of a real photograph in the style of an artist using either a single style sample \cite{gatys2016} or a set of images \cite{sanakoyeu2018styleaware}. Initial work on style transfer by Gatys et al.~\cite{gatys2015neural} proposed a method which exploits a deep CNN (Convolutional Neural Network) pretrained on a large dataset of natural images. 
Their costly computational optimization process has been replaced by an efficient encoder-decoder architecture in recent works \cite{johnson,sanakoyeu2018styleaware,dumoulin2016learned,Babaeizadeh2018AdjustableRS,Gupta2017CharacterizingAI} that efficiently generate the stylized output in a single feed-forward pass. 
While \cite{johnson} has proven that an encoder-decoder architecture is both fast and effective for transferring style, it acts as a black-box model, lacking interpretability and accurate control of style injection: content transformation is performed indirectly, meaning there is no explicit control which part of the network carries out the stylization of photos and to what extend. To address this issue, \cite{sanakoyeu2018styleaware} introduced a fixpoint loss that ensures stylization has converged and reached a fixpoint after one feed-forward pass. This style-aware content loss forces the stylization to take place in the decoder. However, the main issue remains: The decoder alters style, synthesizes the stylized image, and upsamples it. All these individual tasks cannot be learned and controlled individually.\par
As a remedy, we introduce a novel \textit{content transformation block} between encoder and decoder 
allowing control over stylization and achieving a style-aware editing of content images.
We force the encoder to explicitly extract content information; the content transformation block $\mathcal{T}$ then modifies the content information in a manner appropriate to the artist's style. Eventually the decoder superimposes the style on the altered content representation. Our approach measures the content similarity between the content target image and stylized image before and after the transformation.\par 
In contrast to previous work, stylization should be object specific and depending on the underlying object, the style transformation needs to adapt. 
The Cubist style of Picasso, for example, tends to reduce the human nose to a simple triangle or distorts the location of the eyes. Therefore, we further investigate, if we can achieve an object-specific alteration. We utilize similar content appearing in photographs and style samples to learn how style alters content details. We show that by using a prominent, complex, and diverse object class, i.e., persons, our model can learn how details are to be altered in a content-and style-aware manner. Moreover, the model learns to generalize beyond this one particular object class to diverse content. This is crucial to stylize also modern objects like computers which an artist like Monet never painted. 
In addition, we propose a local feature normalization layer to reduce the number of artifacts in stylized images, significantly improving results when moving to other image collections (i.e. from Places365~\cite{zhou2017places} to ImageNet~\cite{ImageNet}) and increasing the image resolution. To validate the performance of our approach, we perform various qualitative and quantitative evaluations of stylized images and also demonstrate the applicability of our method to videos. Additional results can be found on the \href{https://compvis.github.io/content-targeted-style-transfer/}{project page}.



\section{Related Work}
\textbf{Texture synthesis} Neural networks were long used for texture synthesize \cite{gatys2015texture}; feed-forward networks then enable a fast synthesis, however these methods often display a lack of diversity and quality \cite{johnson,ulyanov2016texture}. To circumvent this issue, \cite{li2017diversified} propose a deep generative feed-forward network, which allows to synthesize multiple textures within one single network. \cite{gatys2017controlling} has demonstrated how control over spatial location, color and across spatial scale leads to enhanced stylized images, where regions are altered by different styles; control over style transfer has been extended to stroke sizes \cite{jing2018stroke}. \cite{risser2017stable} used a multiscale synthesis pipeline for spatial control and to improve texture quality and stability.\\
\textbf{Separating content and style} The integration of localized style losses improved the separation of content and style. In order to separate and recombine style and content in an image, works have utilized low-level features for texture transfer and high-level information to represent content using neural networks \cite{gatys2016}. \cite{collomosse2017sketching,bautista2016cliquecnn,patrick_esser,wilber2017bam} focused on distinguishing between different contents, styles and techniques in the latent space; to translate an image to another image is a vision problem, where the mapping between input and output image relies on aligned pairs. To avoid the need for paired examples, \cite{zhu2017unpaired} presented an adversarial loss coupled with a cycle consistency loss to effectively assign two images. On the basis of \cite{zhu2017unpaired}, \cite{sanakoyeu2018styleaware} has proposed an approach, where a style-aware content loss helps to focus on those content details relevant for a style. A combination of generative Markov random field (MRF) models and deep convolutional neural networks have been used for the task of synthesizing content of photographs and artworks \cite{li2016combining}.\\
\textbf{Real-time and super-resolution} The processing time of style transfer and the resolution of images have been further addressed. Scholars aimed to achieve stylization in real time and in super-resolution using an unsupervised training approach, where either neural network features and statistics compute the acquired loss function \cite{johnson} or a multiscale network is employed \cite{ulyanov2016texture}. To achieve a better quality for stylized images in high resolution, \cite{wang2017multimodal} propose a multimodal convolutional network, which performs a hierarchical stylization by utilizing multiple losses of increasing scales.\\
\textbf{Stylizing videos} While these works have approached the task of style transfer for input photographs, others concentrated on transferring artistic style to videos \cite{ruder2016artistic,huang2017real,sanakoyeu2018styleaware,ruder2018artistic}, using feed-forward style transfer networks \cite{chen2017stylebank} or networks, which do not rely on optical flow at test time \cite{huang2017real} to improve the consistency of stylization.   
\section{Approach}
    \begin{figure*}
    \begin{center}
        \includegraphics[width=.8\textwidth,height=.3\textheight]{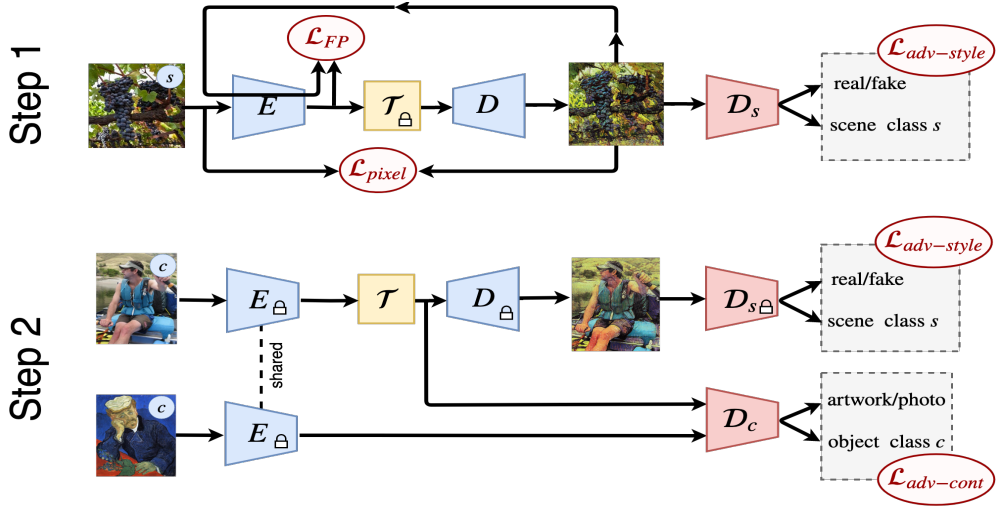}
    \caption{The two figures describe the two alternating training steps. The first step (top) is designated to obtain an artistic stylization while retaining the content information of the input photograph. The second step (bottom) trains the content transformation block $\mathcal{T}$ to alter the image content in a style-specific style. The lock sign indicates that the weights are fixed. See Approach section for further details.}
    \label{fig:pipeline}
    \end{center}   
    \end{figure*}

    \begin{figure}[t!]
    \centering

    \end{figure}
    
Let $\mathbf{Y}$ be a collection of images that defines a style. We extract the very essence of an artistic style presented in $\mathbf{Y}$ and learn to transfer it onto images from a different dataset $\mathbf{X}$, such as photos. This formulation resembles a typical unsupervised image translation problem, which requires a generator $\mathcal{G}$ (usually consisting of the encoder $E$ and decoder $D$) and a discriminator $\mathcal{D}$  trained against each other: one mimics the target distribution $\mathbf{Y}$, the other one distinguishes between the authentic sample $y\in\mathbf{Y}$ and the stylized sample $D(E(x))$ for $x \in \mathbf{X}$.  
Hence, we can extract the style by solving the min-max optimization task for the standard adversarial loss:
\begin{equation}
\label{eq:adv-loss}
\begin{gathered}
\mathcal{L}_{adv} := 
\underset{y\sim \mathbf{Y}}{\mathbb{E}}[log(\mathcal{D}(y))] + \\
\underset{x \sim \mathbf{X}}
{\mathbb{E}}
\left[ log \left(1- \mathcal{D}_s\left(D(E(x))\right)\right) \right ]
\end{gathered}
\end{equation}
Let $s$ be additional content information that is easily available, i.e., we utilize a simple coarse scene label of the image $x$. Now the discriminator should not only discern real from synthesized art. It should also enforce that the scene information is retained in $D(E(x))$ by the stylization process,  
\begin{equation}
\label{eq:cadv-loss}
\begin{gathered}
\mathcal{L}_{cadv} := 
\underset{y\sim \mathbf{Y}}{\mathbb{E}}[log(\mathcal{D}(y))] + \\
\underset{(x, s) \sim \mathbf{X}}
{\mathbb{E}}
\left[ log \left(1- \mathcal{D}\left(D(E(x))|s\right)\right) \right ]
\end{gathered}
\end{equation}
In contrast to a GAN framework that generates an image from a random vector $z$, style transfer not only requires to stylize a real input photograph $x$ but also to retain the \textit{content} of the input image after stylization. The simplest solution would be to enforce a per-pixel similarity between the input $x\sim\mathbf{X}$ and stylized image $\mathcal{G}(x)$:
\begin{equation}
\label{eq:pxl-loss-primary}
\begin{gathered}
\mathcal{L}_{pxl} :=  
\underset{x \sim \mathbf{X}}
{\mathbb{E}} 
[\| D(E(x)) - x\|^2_2].
\end{gathered}
\end{equation}
However, this loss alone would counter the task of stylization, since the image should not be the same afterwards on a per-pixel basis. Previous work \cite{johnson,gatys2015neural} has utilized a pretrained perceptual loss \cite{vgg}. Since this loss is pretrained on an image dataset unrelated to any specific style, it cannot account for the characteristic way in which an artist alters content. Rather, we enforce the stylization to have reached a fixpoint, meaning that another round of stylization should not further alter the content. The resulting fixpoint loss measures the residual in the style-specific encoding space $E(\cdot)$, 
\begin{equation}
\label{eq:FP-loss-primary}
\begin{gathered}
\mathcal{L}_{FP} := 
\underset{x \sim \mathbf{X}}
{\mathbb{E}} 
[\| E(D(E(x))) - E(x)\|^2_2].
\end{gathered}
\end{equation}
\subsection{Content Transformation Block}
While a painting of an artist is associated with one style, it is noticeable that style affects image regions differently: to emphasize the importance of an individual object, artists would use a more expressive brushstroke or deform it to a higher degree. Therefore, we do not only want to learn a simple stylization but a content-specific stylization. Thus each content detail must be stylized in a manner specific to this particular content category. This means that a stylized human figure should resemble how an artist has painted the figure in a specific style and not an arbitrary object, such as a vase or a chair. We enforce this capability by pulling images of similar content – but from different domains(art and photograph) –  closer to each other in the latent space, while keeping dissimilar content images apart from each other. To be more specific, we force the content representation of an input photograph belonging to a specific class $c$ to become more similar to the input painting's content representation of the same class. 
To achieve this, we introduce a content transformation block $\mathcal{T}$ transforming the output representation of the encoder $E$. We train this block in the adversarial fashion: the discriminator $\mathcal{D}_c$ has to distinguish between the representation of the real artworks' content and the transformed representation of the input photographs. But since we strive to obtain a content specific stylization, the discriminator $\mathcal{D}_c$ also has to classify the content class $c$ of the artwork $y$ and the content class of the input photograph $x$. Supplied with the content information $c$ discriminator becomes more sensitive to content specific visual clues and enforces the content transformation block to mimic them in an artistic way. 
\begin{equation}
\label{eq:cadv-cont-loss}
\begin{gathered}
\mathcal{L}_{adv-cont} := 
\underset{(y, c)\sim \mathbf{Y}}{\mathbb{E}}[log(\mathcal{D}_c(E(y)|c)))] + \\
\underset{(x, c) \sim \mathbf{X}}
{\mathbb{E}}
\left[ log \left(1- \mathcal{D}_c\left(\mathcal{T}(E(x))|c\right)\right) \right ]
\end{gathered}
\end{equation}

In terms of neural architecture the $\mathcal{T}$ represents a concatenation of nine ``residual blocks''. Each block consists of six consecutive blocks with a skip connection:   \texttt{conv}-layer, \texttt{LFN}-layer, \texttt{lrelu}-activation, \texttt{conv}-layer, \texttt{LFN}-layer, \texttt{lrelu}-activation.

\subsection{Local Feature Normalization Layer}
Many approaches using convolutional networks for image synthesis suffer from domain change (i.e. from photos of landscapes to faces) or synthesis resolution change. As a result, the inference size is often identical to the training size or the visual quality of the results deteriorates when switching to another domain. Reason being that instance normalization layers overfit to image statistics and the layer is not able to generalize to another image. We can improve the ability to generalize by enforcing stronger normalization through our local feature normalization layer. This layer normalizes the input tensor across a group of channels and also acts locally, not seeing the whole tensor but only the vicinity of the spatial location. 
Formally, for an input tensor $T\in\mathbb{R}^{B\times H\times W\times C}$, where $B$ stands for the samples number, height $H$, width $W$ and having $C$ channels, we can define a \textit{Local Feature Normalization Layer(LFN)} with parameters $WS$ denoting spatial resolution of the normalization window and $G$ - number of channels across which we normalize: $$LFN(\cdot | WS, G): \mathbb{R}^{B\times H\times W\times C} \longrightarrow \mathbb{R}^{B\times H\times W\times C}.$$ To simplify the notation, we first define a subset of the tensor $T$ around $(b, h, w, c)$ with a spatial window of size $WS \times WS$ and across a group of $G$ neighbouring channels:
\begin{multline}
  \mathbb{B}_{WS, G}(T, b, h, w, c):=  \\
  \left\{T(b,x,y,z) \left\|
        \begin{array}{ccc}
        h-WS/2 &\leq x \leq& h+WS/2 \\
        w-WS/2 &\leq y \leq& w+WS/2 \\
        \floor*{\frac{c}{G}}G &\leq z \leq& \floor*{\frac{c}{G}}G+G 
        \end{array}  
  \right\}\right. .
\end{multline}
Finally, we can write out the expression for the Local Feature Normalization Layer applied to tensor $T$ as:
\begin{multline}
LFN(T | WS, G)(b, h, w, c) := \\
\gamma_c 
\frac{T(b, h, w, c) - \textbf{mean}[\mathbb{B}_{WS, G}(T, b, h, w, c)]}{\textbf{std}[\mathbb{B}_{WS, G}(T, b, h, w, c)]} + \beta_c.
\end{multline}
In this equation, similar to the Instance Normalization Layer \cite{instancenorm}, parameters $\gamma,\ \beta\in\mathbb{R}^C$ denote vectors of trainable parameters and represent how to scale and shift each channel; those are learned jointly with other weights of the network via back-propagation.  
However, in practice the computation of $\textbf{mean}$ and $\textbf{std}$ of a large tensor could be a laborious task, so we compute these values only at the selected locations $(b, h, w, c)$ and interpolate for others.

\subsection{Training Details}
The training dataset $\mathbf{X}$ is the union of the Places365 dataset~\cite{zhou2017places} and the COCO dataset~\cite{COCO}, such that for a tuple $(x, c, s)\in\mathbf{X}$ where $x$ is a photograph, $s$ is a  scene class if $x$ is from the Places dataset and $c$ is a content class if $x$ is from the COCO dataset. 
The second dataset $\mathbf{Y}$ contains tuples  $(y, c)$ where $y$ is the artwork and $c$ is the content class. We focus on the content classes ``person''  and a negative class ``non-person''. 
The generator network consists of encoder $E$, transformer block $\mathcal{T}$ and decoder $D$. We utilize two conditional discriminators $\mD_s$ and $\mD_c$ - the former is applied to the input images and stylized outputs. The latter is applied to the content representation obtained by encoder $E$. 
Given this notation the losses become

\begin{equation}
\label{eq:cadv-style-loss}
\begin{gathered}
\mathcal{L}_{adv-style} := 
\underset{(y, c)\sim \mathbf{Y}}{\mathbb{E}}[log(\mathcal{D}_s(y))] + \\
\underset{(x, c, s) \sim \mathbf{X}}
{\mathbb{E}}
\left[ log \left(1- \mathcal{D}_s\left(D(\mathcal{T}(E(x)))|s\right)\right) \right ]
\end{gathered}
\end{equation}

\begin{equation}
\label{eq:cadv-cont-loss}
\begin{gathered}
\mathcal{L}_{adv-cont} := 
\underset{(y, c)\sim \mathbf{Y}}{\mathbb{E}}[log(\mathcal{D}_c(E(y)|c)))] + \\
\underset{(x, c, s) \sim \mathbf{X}}
{\mathbb{E}}
\left[ log \left(1- \mathcal{D}_c\left(\mathcal{T}(E(x))|c\right)\right) \right ]
\end{gathered}
\end{equation}

\begin{equation}
\label{eq:FP-loss}
\begin{gathered}
\mathcal{L}_{FP} := 
\underset{(x, c, s) \sim \mathbf{X}}
{\mathbb{E}} 
[\| E(D(\mathcal{T}(E(x)))) - E(x)\|^2_2].
\end{gathered}
\end{equation}

\textbf{Training procedure}
For variables $\theta_E, \theta_D, \theta_\mathcal{T}, \theta_{\mathcal{D}_c}, \theta_{\mathcal{D}_s}$ denoting parameters of the blocks $E, D, \mathcal{T},\mathcal{D}_c, \mathcal{D}_s$. Training is performed in two alternating optimization steps. 

The first step designated to obtain an accurate content extraction in encoder $E$ and to learn a convincing style injection by decoder $D$.
\begin{equation}
\label{eq:objective-step1}
\begin{gathered}
\underset{\theta_E, \theta_D}{\text{min}}\ \underset{\theta_{\mathcal{D}_s}}{\text{max}}\ 
\lambda_{\mathcal{L}_{pxl}}\mathcal{L}_{pxl} + \\ 
\lambda_{\mathcal{L}_{FP}}\mathcal{L}_{FP} +
\lambda_{\mathcal{L}_{adv-style}}\mathcal{L}_{adv-style}
\end{gathered}
\end{equation}

The second step is aimed to learn style-specific content editing by the block $\mathcal{T}$.

\begin{equation}
\label{eq:objective-step1}
\begin{gathered}
\underset{\theta_\mathcal{T}}{\text{min}}\ \underset{\theta_{\mathcal{D}_c}}{\text{max}}\ 
\lambda_{\mathcal{L}_{adv-cont}}\mathcal{L}_{adv-cont} + \\
\lambda_{\mathcal{L}_{adv-style}}\mathcal{L}_{adv-style}
\end{gathered}
\end{equation}

Please see Figure \ref{fig:pipeline} illustrating the alternating steps of the training.
    
\section{Experiments and Discussion}
\subsection{Stylization Assessment}
\begin{figure}[t!]
    \centering
    \includegraphics[width=\linewidth,height=\linewidth]{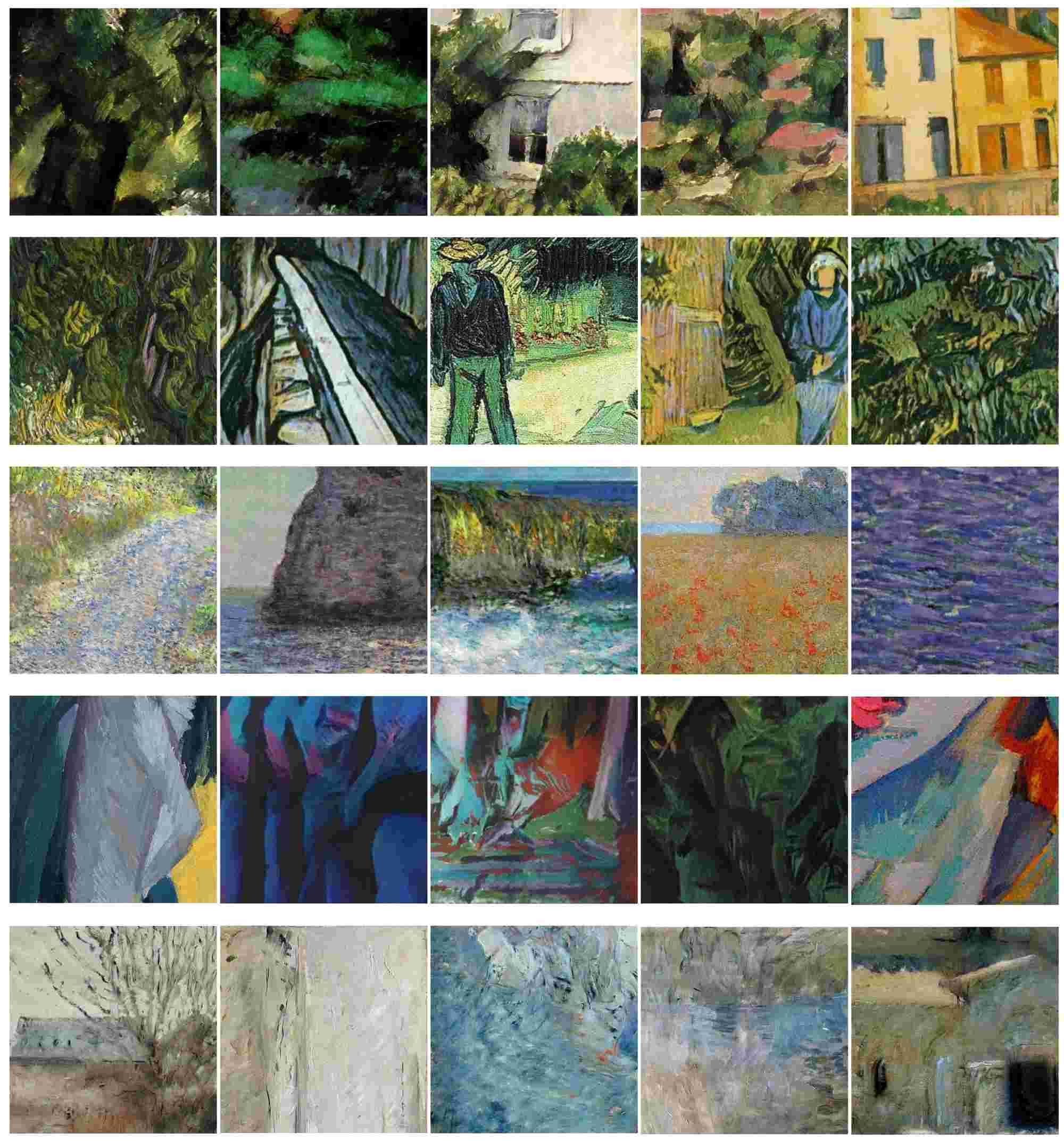}
    \caption{Can you guess which patches are real and which were generated by our approach? 
    Each row contains three patches generated by our model and two real patches. Artists: (from the top) Cezanne, van Gogh, Claude Monet, Ernst Ludwig Kirchner and Berthe Morisot. The solution is provided on the last page.}
    \label{fig:patches_puzzle}
\end{figure}
\begin{figure}
    \centering
    \includegraphics[width=0.5\textwidth,height=0.81\textwidth]{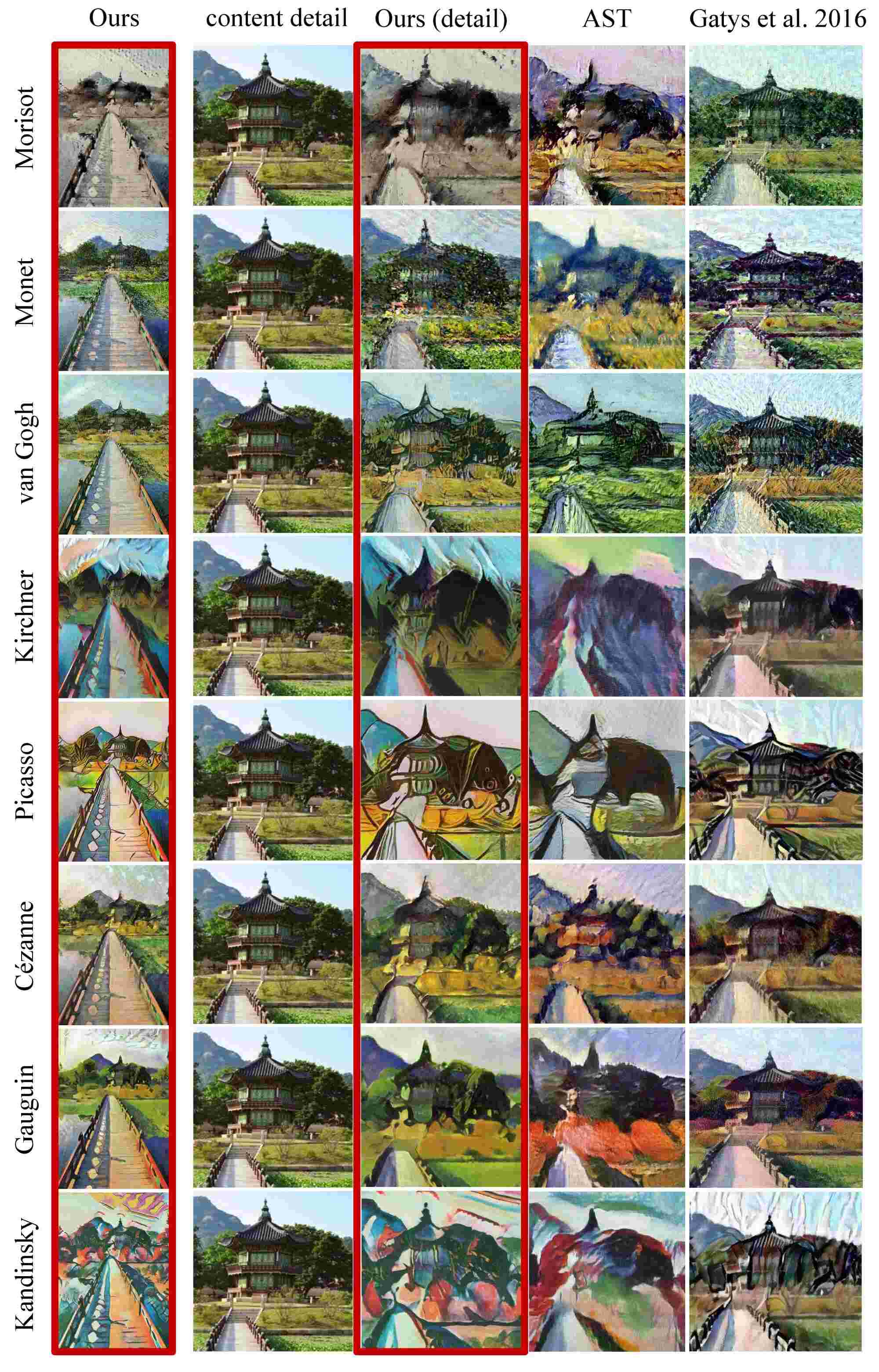}
    \caption{Qualitative comparison: the first column shows the entire image stylized by our approach (the real content image is provided on top), a detailed view is presented in the second. The third column shows the real photo detail respectively, while the last two display results obtained by other methods. Zoom in for a better view and details.}
    \label{fig:content_retention}
\end{figure}

To measure the quality of the generated stylizations we provide qualitative results of our approach and perform several quantitative experiments which we describe below.\\
\textbf{Deception rate.} This metric was introduced in \cite{sanakoyeu2018styleaware} to asses how good the target style characteristics are preserved in the generated stylizations. 
A network pre-trained for artist classification should predict the artist which was used to generate the stylization. The deception rate is then calculated as the fraction of times the network predicted the correct artist. We report the deception rate for our and competing methods in Tab.~\ref{tab:score_user_study} in the first column, where we can see that our approach outperforms other methods by a significant margin.\\
\textbf{Expert and non-expert score.}
We also perform human evaluation studies to highlight the quality of our stylization results.
Given a content image patch, we stylize it with different methods and show results alongside a patch from a real painting to experts and non-experts. Both are asked to guess which one of the shown patches is real. The score is the fraction of times the stylization generated by this method was selected as the real patch.
This experiment is performed with experts from art history and people without art education. Results are reported in Tab.~\ref{tab:score_user_study}.\\
\textbf{Expert preference score.}
In addition, we asked art historians to choose which 
of the stylized images resemble the style of the target artist the most.
Then the expert preference score (see Tab.~\ref{tab:score_user_study}) is calculated as the fraction of times the stylizations of the method was selected as the best among the others. The quantitative results in Tab.~\ref{tab:score_user_study} show that both experts and non-experts prefer our stylizations in comparison to images obtained by other methods. \\
\textbf{Content retention evaluation.}
To quantify how well the content of the original image is preserved, we stylize the ImageNet \cite{ImageNet} validation dataset with different methods and compute the accuracy using pretrained VGG-16 \cite{vgg} and ResNet-152 \cite{resnet} networks averaged across $8$ artists. Results presented in Tab.~\ref{tab:classification} show that the best classification score is achieved on stylizations by CycleGAN~\cite{cyclegan} and Gatys et al.~\cite{gatys2016}, since both methods barely alter the content of the image. However, our main contribution is that we significantly outperform the state-of-the-art AST~\cite{sanakoyeu2018styleaware} model on the content preservation task, while still providing more convincing stylization results, measured by the deception rate in Tab.~\ref{tab:score_user_study}.
\\
\textbf{Qualitative comparison.} We compare our method qualitatively with existing approaches in Fig.~\ref{fig:content_retention}. The reader may also try to guess between real and fake patches generated by our model in Fig.\ref{fig:patches_puzzle}. More qualitative comparisons between our approach and other methods are available in the supplementary material.

\begin{table}
    \begin{center}
    \scriptsize
    \begin{tabular}{lc||ccc}
    \hline
    Method & Deception & Non-Expert  & Expert     &  Expert \\
          & rate~\cite{sanakoyeu2018styleaware}   & deception  & deception & preference \\
          & & score & score & score \\
    \hline

    \xrowht{8pt}AdaIn~\cite{adain}             & 0.067 & 0.033 & 0.016 & 0.019\\
    WCT~\cite{universal_style}     & 0.030 & 0.033 & 0.001 & 0.009\\
    PatchBased\cite{pb}            & 0.061 & 0.118 & 0.011 & 0.038\\
    Johnson et al.~\cite{johnson}  & 0.087 & 0.013 & 0.001 & 0.010\\ 
    CycleGan~\cite{cyclegan}       & 0.140 & 0.026 & 0.031 & 0.010\\
    Gatys et al.~\cite{gatys2016}  & 0.221 & 0.088 & 0.068 & 0.118\\
    AST \cite{sanakoyeu2018styleaware}                   
                                  & 0.459 & 0.056 & 0.131 & 0.341\\
    \textbf{Ours}                  & \textbf{0.582} & \textbf{0.178} & \textbf{0.220} & \textbf{0.456} \\
    \hline\hline
    \xrowht{8pt}Wikiart test      & 0.6156 & 0.454 & 0.528 & - \\
    Photos & 0.002 & - & - & -\\
    \hline      
    \end{tabular}
    \end{center}
    
    \caption{A higher score indicates better stylization results. All scores are averaged over $8$ different styles. The row "Wikiart test"~\cite{karayev2013recognizing} shows accuracy on real artworks from the test set. The deception rate for "Photos" shows how often photos were missclassified by the network as real paintings of the target artist.}
    \label{tab:score_user_study}
\end{table}

\begin{table}
    \begin{center}
        \scriptsize
        \begin{tabular}{cc||cc||cc}
        \hline
        \xrowht{5pt}Network & Original & \textbf{Ours} &  AST & Gatys & CycleGAN \\
        & photos & & \cite{sanakoyeu2018styleaware} & \cite{gatys2016} & \cite{cyclegan} \\[0.5ex]
        \hline
        \xrowht{8pt}VGG-16 & 0.710  & \textbf{0.016} & 0.009 & 0.271 & 0.198 \\
                ResNet-152 & 0.783	& \textbf{0.057} & 0.032 & 0.389 & 0.341 \\
        \hline      
     
        \end{tabular}
     \end{center}
    \caption{Top-1 classification accuracy on stylized images from validations set of ImageNet \cite{ImageNet} using the networks pretrained on ImageNet. Note that the classification accuracy of our model is higher then the state-of-the art model~\cite{sanakoyeu2018styleaware}. We include results for \cite{gatys2016} and \cite{cyclegan}, but they are not directly comparable to our method since they barely alter the content of the input image. In the second column we present the classification accuracy on the input photos.}
  \label{tab:classification}
\end{table}

\begin{table}
    \begin{center}
        \scriptsize
        \begin{tabular}{lc||ccc}
        \hline
        \xrowht{5pt}Feature  & Photographs & \textbf{Ours} & Ours & AST~\cite{sanakoyeu2018styleaware} \\
        extractor & & & w/o $\mathcal{T}$&  \\[0.5ex]
        \hline
        \xrowht{8pt}VGG-16 &  1.108  &	\textbf{0.756} &	0.882 &	0.812 \\
        VGG-19 &  1.025	 &  \textbf{0.724} &	0.838 & 0.808 \\
        \hline      
        \end{tabular}
    \end{center}
    \caption{The table summarizes RSSCD computed using different classification networks for different stylization methods. The score characterizes the content dissimilarity between real artworks and stylized images, relative to the average content dissimilarity between the artworks.  The lower the better.}
    \label{tab:SSCS}
\end{table}


        

\begin{table}
    \scriptsize
    \centering
    \begin{tabular}{cc||ccc}
    \hline
    \xrowht{5pt} & Photographs & \textbf{Ours} & Ours & AST~\cite{sanakoyeu2018styleaware} \\
     & & & w/o $\mathcal{T}$ &   \\[0.5ex]
    \hline

    \xrowht{8pt}Accuracy & 0.953 & \textbf{0.659} & 0.598 & 0.548 \\
    Recall & 0.978 & \textbf{0.375} & 0.209 & 0.109 \\
    Precision & 0.927 & 0.864 & \textbf{0.937} & 0.893 \\
    F1-Score & 0.954 & \textbf{0.524} & 0.343 & 0.195 \\
    \hline      
    \noalign{\smallskip}
    \end{tabular}
    \caption{Results of the person detection on the stylized images from COCO dataset \cite{COCO} using Mask-RCNN~\cite{matterport_maskrcnn_2017}. Columns from left to right: Person detection on photos; on stylized images by our method; on stylized images by our method without transformation block; on stylized images by AST~\cite{sanakoyeu2018styleaware}.}

    \label{tab:coco_clsf}

\end{table}
\subsection{Ablation Study}
\subsubsection{Content Transformation}

\begin{figure}[t!]
    \centering
    \includegraphics[width=0.4\textwidth,height=0.4\textwidth]{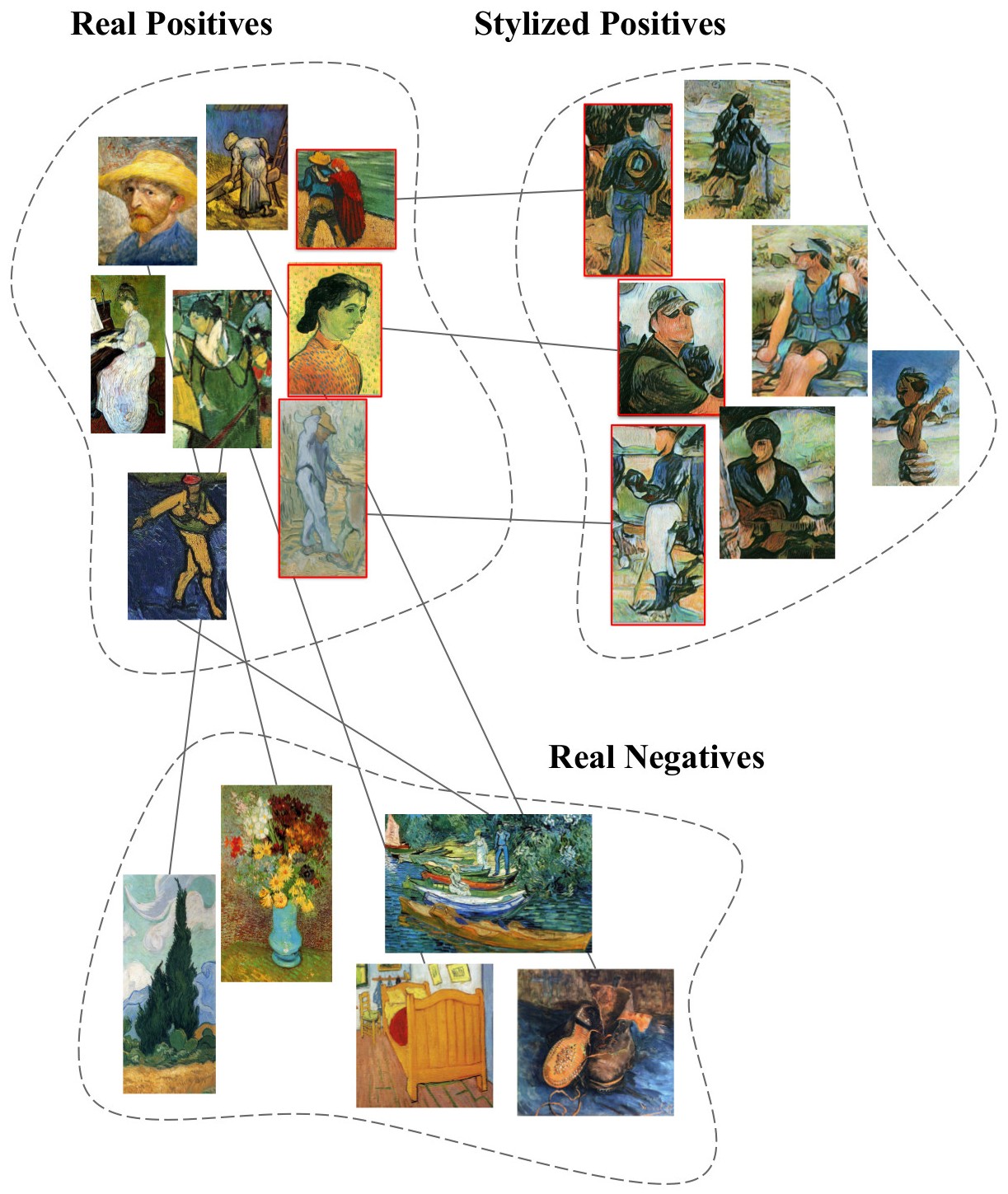}
    \caption{Illustration of the RSSCD measure. Real positive, real negative and stylized positive images are mapped into the feature space using deep CNN.  We then compute the average distance from a stylized positive to the closest real positive and divide it by the average distance between real positives and negatives. Positive images correspond to the class ``person''.}
    \label{fig:SSCS}
\end{figure}

\textbf{Relative style-specific content distance.}
To verify that the image content is transformed in a style-specific manner,
we introduce a quantitative measure, called \textit{relative style-specific content distance} (RSSCD).
It measures the ratio between the average distance of the generated image stylizations to the closest artworks and the average distance between all the artworks. 
Distances are computed using the features $\phi(\cdot)$ of the classification CNN pretrained on ImageNet.
Then, RSSCD is defined as 
$$ \text{RSSCD} := \frac{\frac{1}{|Z_{p}|}\sum\limits_{z \in Z_{p}} \underset{y \in Y_{p}}{\text{min}}\|\phi(z) - \phi(y)\|_2} {\frac{1}
{|Y_{p}||Y_{n}|}
\sum\limits_{y_p \in Y_{p}, y_n \in Y_{n}}
\| \phi(y_p) - \phi(y_n) \|_2},$$
$Z_p$ denotes the set of stylizations of the positive content class (e.g., person), $Y_p$ denotes the set artworks of the positive content class, and $Y_n$ denotes all other artworks
(see Fig.~\ref{fig:SSCS} for an illustration).

We report the RSSCD for our model with and without $\mathcal{T}$. For comparison we also evaluate the state-of-the-art approach AST~\cite{sanakoyeu2018styleaware}. Here, we use class "person" as the positive content class and two pretrained networks as content feature extractors $\phi(\cdot)$, namely VGG-16 and VGG-19~\cite{vgg}.
As can be seen in Tab.~\ref{tab:SSCS}, the content transformation block significantly decreases the distance between the stylized images and original van Gogh paintings, proving its effectiveness.


We measure how well our model retains the information present in the selected ``person'' class and compare it to both the model not using $\mathcal{T}$ and to the AST~\cite{sanakoyeu2018styleaware}. We run the Mask-RCNN detector~\cite{matterport_maskrcnn_2017} on images from the COCO~\cite{COCO} dataset stylized by different methods and compute the accuracy, precision, recall and F1-score. From results ins Tab.~\ref{tab:coco_clsf} we conclude that the proposed block $\mathcal{T}$ helps to retain visual details relevant for the ``person'' class.

\begin{figure}[t!]
    \centering 
    
    \includegraphics[width=0.45\textwidth,height=1.\textwidth]{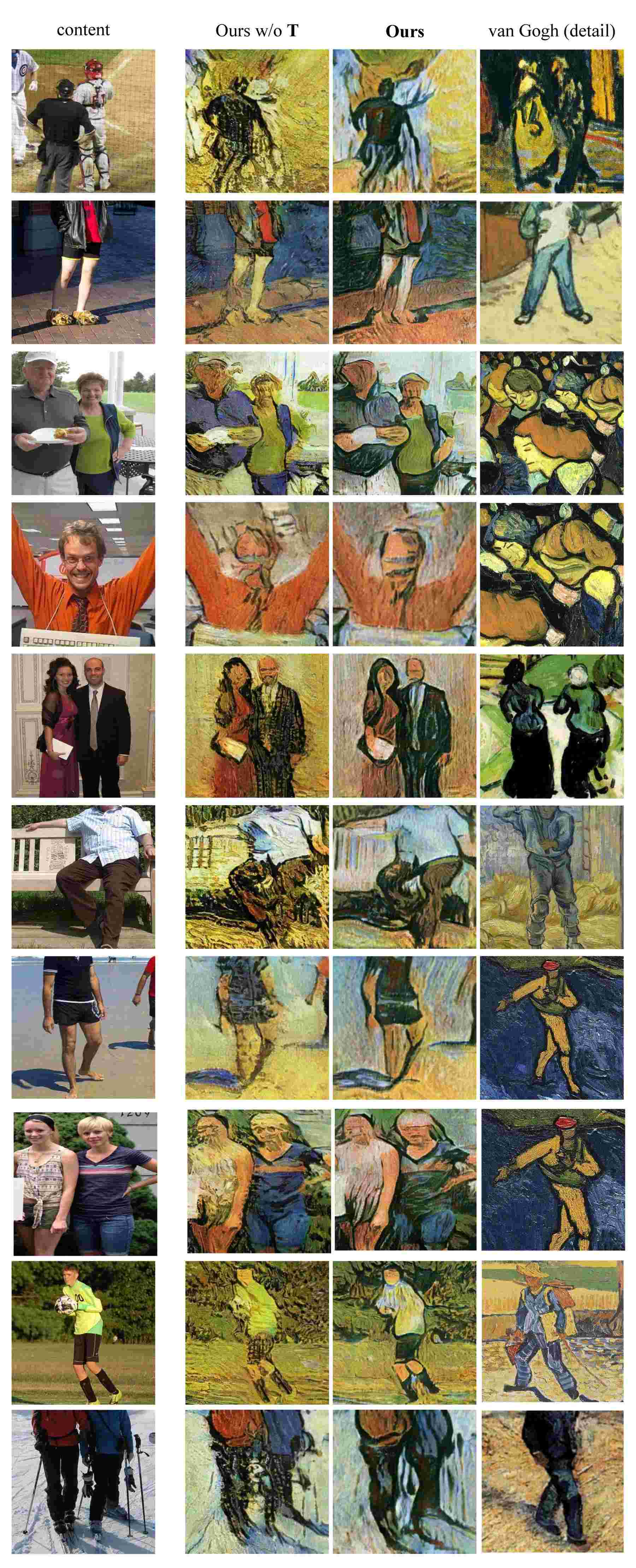}
    \caption{Shows the impact of the content transformation block: column two shows results without using a content transformation block. The third displays identical details with a content transformation block, emphasizing outlines of figures in a manner appropriate to van Gogh's style. The last column provides details from the artist's paintings to highlight the validity of our approach. Best seen on screen and zoomed in.}
    \label{fig:content_transformation}
\end{figure}

In Fig.~\ref{fig:content_transformation} we show stylizations of our method
with and without content transformation block. We recognize that applying the content transformation block alters the shape of the human figures in a manner appropriate to van Gogh’s style resulting in curved forms (cf. the crop-outs from original paintings by van Gogh provided
in the 4th column of Fig.~\ref{fig:content_transformation}).
For small persons, the artist preferred to paint homogeneous regions with very little texture. This is apparent, for example, in the stylized patches in row one and six. 
Lastly, 
while van Gogh's self-portraits display detailed facial features, in small human figures he tended to remove them (see our stylizations in 3rd and 4th rows of Fig.~\ref{fig:content_transformation}). This might be due to his abstract style, which included a fast-applied and coarse brushstroke.

\subsubsection{Generalization Ability}
The transformer block $\mathcal{T}$ learns to transform content representation of the photographs of class ``person'' in such a way that it becomes indistinguishable of the  content representation of artworks of the same class. 
Though transformation has been learned for one only class ``person'' it can still generalize to other classes. To measure this generalization ability we compute the deception rate\cite{sanakoyeu2018styleaware} and non-expert deception scores on stylized patches for classes ``person'' and non-``person'' separately. The evaluation results are provided in Tab.~\ref{tab:stylization_generalization} and indicate improvement of the stylization quality for unseen content.
\begin{table}
    \begin{center}
    \scriptsize
    \begin{tabular}{l||cc|cc}
    \hline
    Method & Deception & Deception & Non-Expert & Non-Expert  \\
                       & rate~\cite{sanakoyeu2018styleaware}   &  rate  & deception score & deception score  \\
          & ``person'' & non-``person'' & ``person'' & non-``person''\\
    \hline
    AST                    & 0.398 & 0.485 & 0.016 & 0.086 \\
    Ours w/o $\mathcal{T}$ & 0.521 & 0.541 & 0.127 & 0.143 \\
    \textbf{Ours}          & \textbf{0.618} & \textbf{0.563} & \textbf{0.210} & \textbf{0.165} \\
    
    \hline\noalign{\smallskip}
    
    \end{tabular}
    \end{center}
    
    \caption{Stylization quality for different content classes. Our model has significantly improved stylization quality compared to the state-of-the-art AST\cite{sanakoyeu2018styleaware} model. The $\mathcal{T}$ block improves the deception rate and preference score on both classes. The higher the better.}
    \label{tab:stylization_generalization}
\end{table}

\subsubsection{Artifacts Removal}
\begin{figure}[t!]
    \centering
    \includegraphics[width=0.4\textwidth,keepaspectratio]{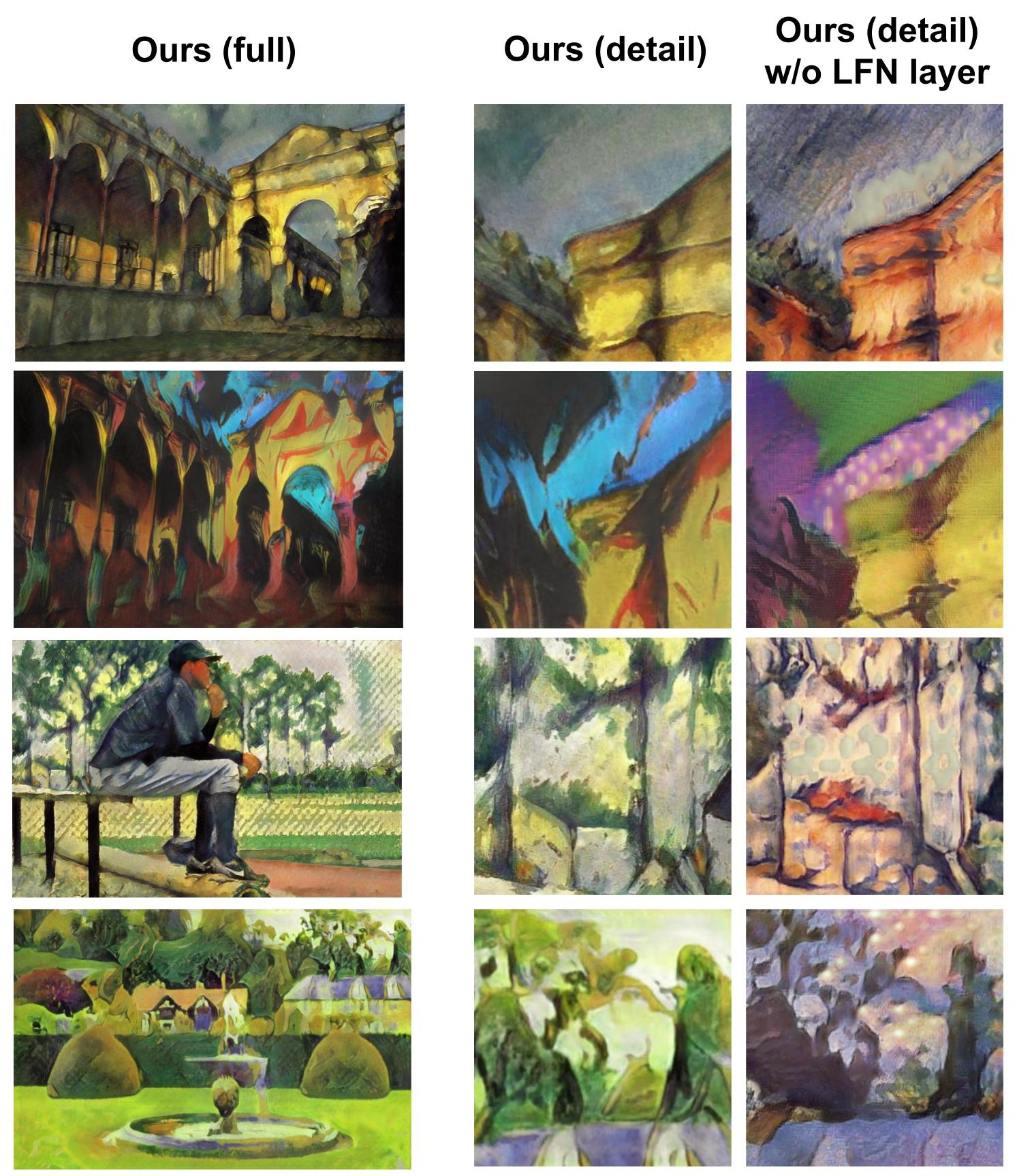}
    \caption{From left to right: stylized image with LFN layer, detail of the image, same region for the model without LFN layer. Styles from top to bottom: Cezanne, Kirchner, Cezanne, Paul Gauguin. Local feature normalization significantly reduces the number of artifacts by normalizing the network activation statistics. }
    \label{fig:artifacts}
\end{figure}
To verify the effectiveness of the local feature normalization layer (LFN layer), we perform a visual inspection of learned models and notice prominent artifacts illustrated in Fig.~\ref{fig:artifacts}. We can observe that especially for plain regions with little structure, the model without a LFN layer often produces unwanted artifacts. In comparison, results obtained with an LFN layer show no artifacts in the same regions. 
Notably, for a  model without an LFN layer the number of artifacts increases proportionally to the resolution of the stylized image.

\section{Conclusion}
We introduced a novel content-transformation block designed as a dedicated part of the network to alter an object in a content-and style-specific manner. 
We utilize objects from the same class in content and style target images to learn how content details need to be transformed. Experiments show that from only one complex object category, our model learns how to stylize details of content in general and thus improves the stylization quality for other objects as well.
In addition, we proposed a local feature normalization layer, which significantly reduces the number of artifacts in stylized images, especially when increasing the image resolution or applying our model to previously unseen image types (photos of faces, road scenes etc.). The experimental evaluation showed that both art experts and persons without specific art education preferred our method to others. 
Our model outperforms existing state-of-the-art methods in terms of stylization quality in both objective and subjective evaluations, also enabling a real-time and high definition stylization of videos.\par

\section*{Acknowledgements}
This work has been supported by a hardware donation from NVIDIA Corporation.

\let\thefootnote\relax\footnotetext{Solution to Figure \ref{fig:patches_puzzle}: \\
Cezanne: fake, real, fake, fake, real\\
van Gogh:  real, fake, real, fake, fake \\
Monet: fake, real, fake, real, fake \\
Kirchner: real, fake, fake, fake, real \\
Morisot: fake, real, fake, real, fake.}
\FloatBarrier
\clearpage

{\small
\bibliographystyle{ieee}
\bibliography{egbib}
}

\clearpage


\twocolumn[{%
 \centering
 \LARGE - Supplementary Material - \\[1.5em]
 
 \normalsize
}]

\maketitle

\section{Additional Visual Comparison}
\begin{figure*}
    \begin{center}
    \includegraphics[width=1.0\textwidth,height=1.0\textwidth]{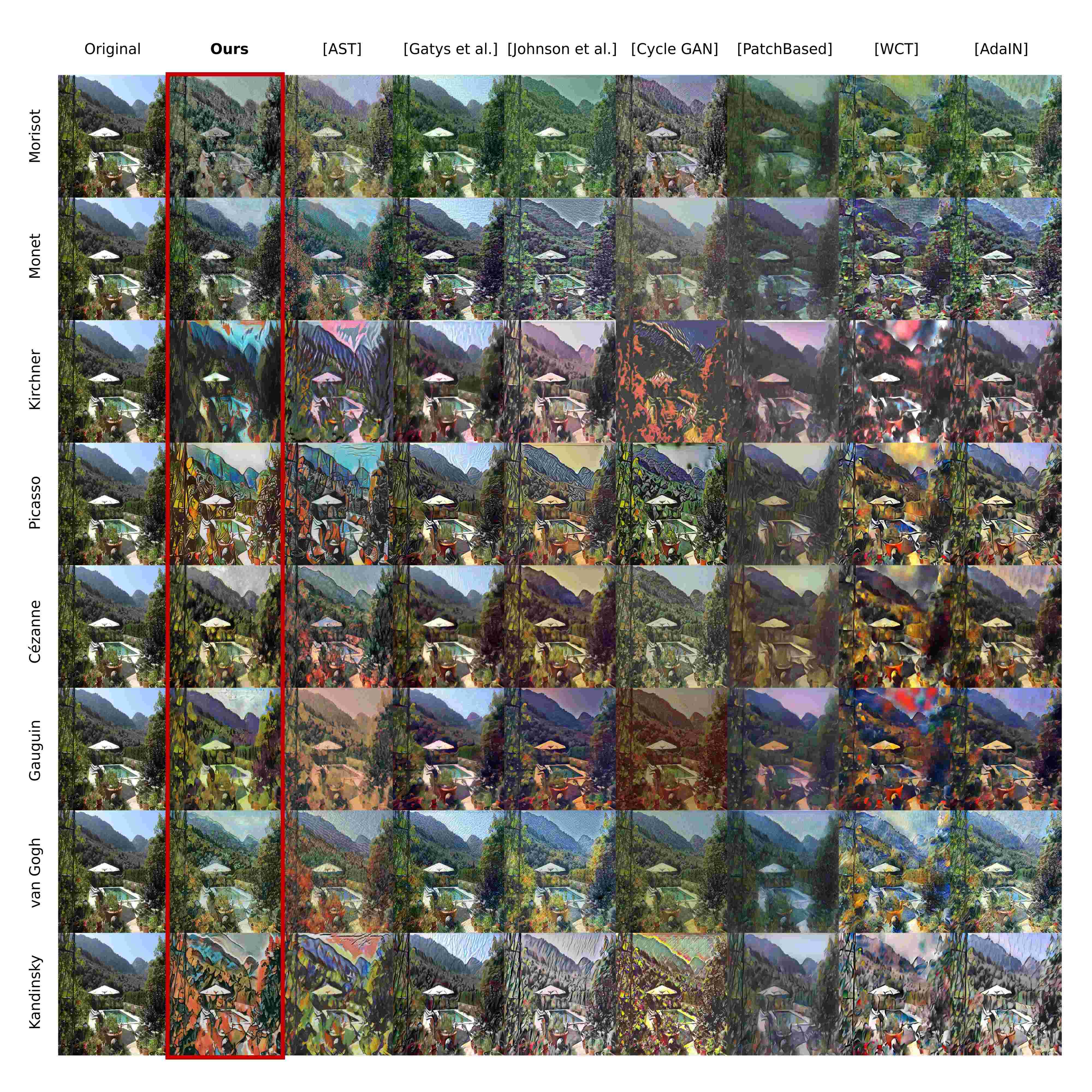}
    \caption{Comparison between our and other methods on full images with the same content for different artists.}
    \label{fig:table_comparison_full_3913}
    \end{center}
\end{figure*}

\begin{figure*}
    \begin{center}0
    \includegraphics[width=1.0\textwidth,height=1.0\textwidth]{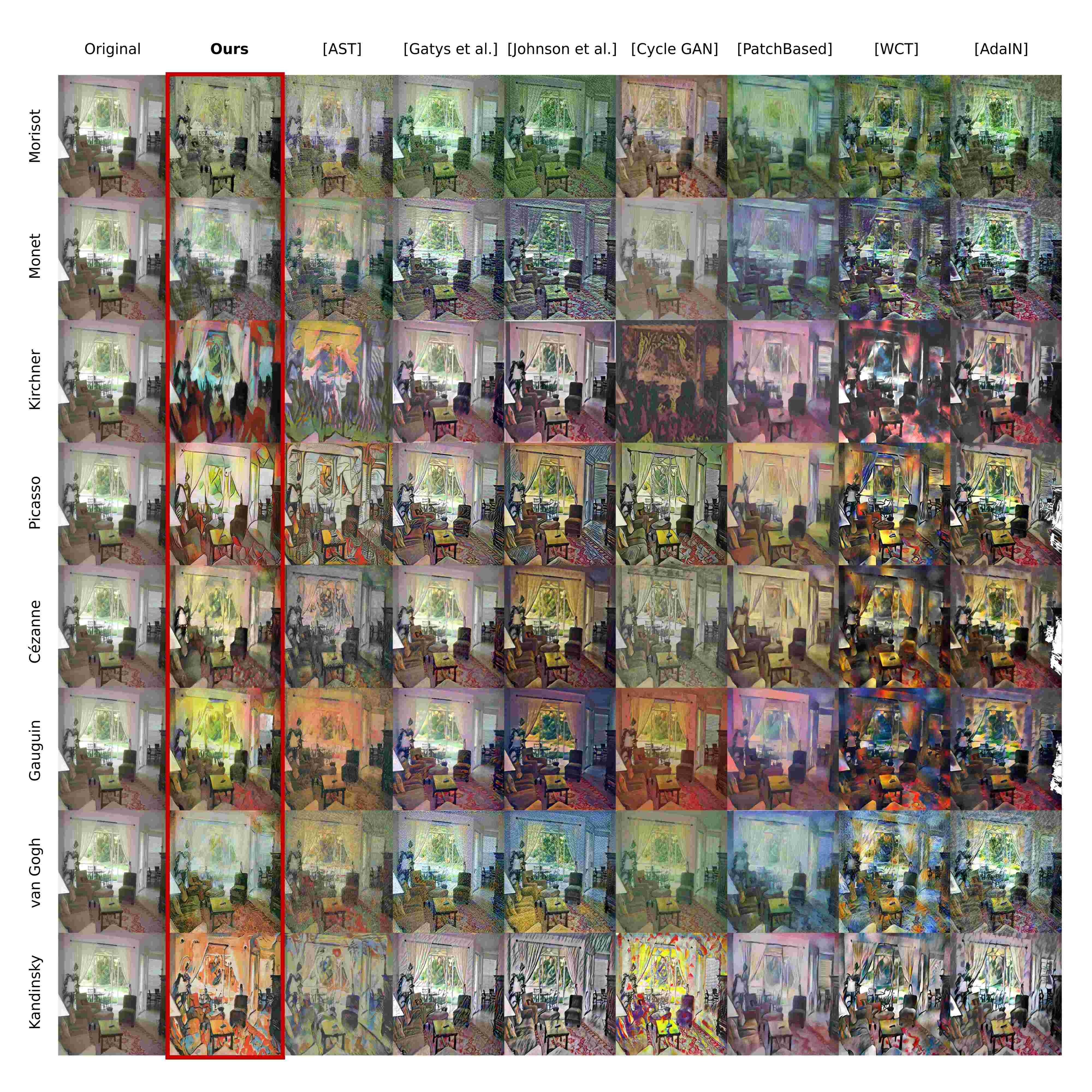}
    \caption{Comparison between our and other methods on full images with the same content for different artists.}
    \label{fig:table_comparison_full_10478}
    \end{center}
\end{figure*}

In this supplementary material, we present additional comparisons with existing style transfer methods for the following $8$ artists: Berthe Morisot, Claude Monet, Ernst Ludwig Kirchner, Pablo Picasso, Paul Cezanne, Paul Gauguin, Vincent van Gogh, and Wassily Kandinsky. Comparisons are presented in Fig.~\ref{fig:table_comparison_full_3913}
and Fig.~\ref{fig:table_comparison_full_10478}. We observe that while providing better stylization than the state-of-the-art AST \cite{sanakoyeu2018styleaware} method, we also retain the content of images better and produce no artifacts; please zoom in for details. All results are generated in resolution with $1280$ pixels as the minimal side of the image. \par
We also stylized two random videos from the internet to show that our method is able to produce real-time high definition stylization of videos, also free of flickering. As input we took two fragments, each $3$ minutes long, from the video \href{https://www.youtube.com/watch?v=9AZ7_yC_t_Q}{Provence: Legendary Light, Wind, and Wine} at timepoints 7:02 and 10:10 and one entire video \href{https://www.youtube.com/watch?v=HPSK4zZtzLI}{Chaplin Modern Times-Factory Scene (late afternoon)}. For best viewing experience, please watch all videos in 4K resolution, since the quality drops significantly due to YouTube's compression algorithms otherwise. For video stylization, we provide a comparison between our method and the AST \cite{sanakoyeu2018styleaware}. In addition, to visualize the necessity of the Content Transformation Block $\mathcal{T}$, we run a side-by-side stylization of our model with and without extra training of $\mathcal{T}$ block. We notice that there is a difference in the way how content is retained and also how parts of the image are highlighted. Our model with block $\mathcal{T}$ achieves better preservation of human figures, especially at smaller scale. The links to the playlists: \href{https://www.youtube.com/playlist?list=PLhronotcCkVlSLxsQTOMhquB2jYEwKCHL}{1}, fragment \href{https://www.youtube.com/playlist?list=PLhronotcCkVlGFRPUmimYqhP6V3ne905q}{2} and fragment \href{https://www.youtube.com/playlist?list=PLhronotcCkVlyieXSc2TBZpdqo2xQ3J4U}{3}.

\section{Implementation Details}

\subsection{Network Architecture Notation}
Our generator network consists of three consequent blocks: encoder $E$, content transformation block $\mathcal{T}$ and decoder $D$. Besides that, we have two discriminators:  $\mathcal{D}_c$ and $\mathcal{D}_s$. For brevity we use the following naming conventions:
\begin{itemize}[*,topsep=0pt]
    \item \texttt{conv-$k\times k$-stride-$s$} denotes a convolutional layer with kernel size $k\times k$ and stride $s$;
    \item \texttt{LFN-G-$g$-W-$w$} denotes a Local Feature Normalization Layer group size $g$ and window size $w$;
    \item \texttt{upscale-$3\times3$} denotes an upscaling layer that consists of nearest neighbor; upscaling is done by a factor of 2, followed by a convolutional layer with kernel size $3 \times 3$ and stride $1$.
    \item \texttt{ResBlock-$3\times3$} denotes a residual block that consists of two convolutional layers with kernel size $3 \times 3$ and stride $1$ followed by \texttt{LFN-G-32-W-32};
    \item \texttt{c$f$s$s$-k-LFN-$g$-$w$-LReLU} denote a $f \times f$ convolution with stride $s$ and $k$ filters followed by \texttt{LFN-G-$g$-W-$w$} layer and LReLU with slope $0.2$ ;

\end{itemize}
All convolutional layers use reflection padding. 

We describe the architecture of the encoder and the decoder in Tab.~\ref{tab:arch_encoder_decoder}.

\begin{table*}
    \centering
    \begin{tabular}{c|c}
    \hline
    Encoder & Decoder  \\
    \hline
    \hline
    Input ($256\times 256 \times 3$ image) & - \\
    \hline
    \texttt{conv-$3\times3$-stride-$1$} & \texttt{conv-$3\times3$-stride-$1$}\\
    \texttt{LFN-G-32-W-128} & (\texttt{ResBlock-$3\times3$}) $\times 9$\\
    
    \texttt{conv-$3\times3$-stride-$2$} & \texttt{upscale-$3\times3$} \\
    \texttt{LFN-G-32-W-64} & \texttt{LFN-G-32-W-16}\\
    
    \texttt{conv-$3\times3$-stride-$2$} & \texttt{upscale-$3\times3$} \\
    \texttt{LFN-G-32-W-32} & \texttt{LFN-G-32-W-32} \\
    
    \texttt{conv-$3\times3$-stride-$2$} & \texttt{upscale-$3\times3$}\\
    \texttt{LFN-G-32-W-32} & \texttt{LFN-G-32-W-64} \\
    
    \texttt{conv-$3\times3$-stride-$2$}  & \texttt{upscale-$3\times3$} \\
    \texttt{LFN-G-32-W-16} & \texttt{LFN-G-32-W-128}  \\
    &  \texttt{conv-$7\times7$-stride-$1$}  \\
    \hline
    & sigmoid \\
    \hline
    \multicolumn{2}{c}{}
    
    \end{tabular}
    \caption{Description of the encoder and the decoder architecture. ReLU layers are omitted for brevity.}
    \label{tab:arch_encoder_decoder}
\end{table*}

\subsubsection{Content Transformation Block $\mathcal{T}$ }
Content transformation block $\mathcal{T}$ consists of $9$ consequent residual blocks \texttt{ResBlock-$3\times3$} with each convolution having $256$ kernels. 

\subsubsection{Architecture of the  Discriminators  $\mathcal{D}_s$ and $\mathcal{D}_c$}
Both discriminators described in Tab.~\ref{tab:arch_discriminator} have a double purpose: predicting the class of the input image and predicting domain of the image (real painting or not). On the one hand, the discriminator $\mathcal{D}_s$ takes images as an input and predicts scene class and domain. The discriminator $\mathcal{D}_c$, on the other hand, takes a feature vector as an input and predicts class (person/non-person) and domain. Both predictions are given as two values in the last line of $\mathcal{D}_c$ architecture in Tab.~\ref{tab:arch_discriminator}. 

For the discriminator $\mathcal{D}_s$ conditions are more complicated: a class of the image scene is predicted in the final layer, see Tab.~\ref{tab:arch_discriminator}. To obtain domain predictions, we attach a convolutional layer of one single kernel to the outputs of the $4$th and the $5$th convolutional layers of the discriminator and compute average mean on its outputs. This value is used as domain prediction.

\begin{table*}
    \begin{center}
    \begin{tabular}{c|c}
    \hline
    $D_C$ & $D_S$  \\
    \hline
    \hline
    Input ($16\times 16 \times 256$ tensor) & Input ($256\times 256 \times 3$ image) \\
    \hline
    \texttt{global avg\ pooling} & 
    \texttt{c$5$s$2$-$128$-LFN-$32$-$128$-LReLU} \\ 
    \texttt{fc-512-ReLU} & \texttt{c$5$s$2$-$128$-LFN-$32$-$64$-LReLU}\\
    \texttt{fc-512-ReLU} & \texttt{c$5$s$2$-$256$-LFN-$32$-$32$-LReLU}\\
    \texttt{fc-2}, \texttt{fc-2} & \texttt{c$5$s$2$-$512$-LFN-$32$-$16$-LReLU}\\
    &
    \texttt{c$5$s$2$-$512$-LFN-$32$-$8$-LReLU}
    \\
    &
    \texttt{c$5$s$2$-$1024$-LFN-$32$-$4$-LReLU}
    \\
    &
    \texttt{c$5$s$2$-$1024$-LFN-$32$-$4$-LReLU}
    \\
    &
    \texttt{conv-$6\times 6$-stride-$1$-LReLU}
    \\
    &
    \texttt{max\_pool}
    \\
    &
    \texttt{fc-num\_classes} \\
    \hline
    
    \end{tabular}
    \end{center}
    \caption{Description of the architecture of discriminators $D_S$ and $D_C$.}
    \label{tab:arch_discriminator}
\end{table*}

\subsection{Training Details}
Training process consists of two stages: a randomly initialized network is trained at first on patches of size $256 \times 256$ pix cropped from the real paintings and patches cropped from the photographs with scene class label for $400000$ iterations with batch size $8$. Afterwards, we continue training procedure on patches of size $768 \times 768$ pix for another $400000$ iterations with batch size $1$ on the two aforementioned datasets. At this stage, we also train on patches of person and non-person class extracted from both paintings and photographs dataset.
At each training stage we use two different Adam \cite{adam} optimizers both with learning rate $2 \times 10^{-4}$: one for discriminators $\mathcal{D}_s,\ \mathcal{D}_c$  and another for encoder $E$, transformation block $\mathcal{T}$ and decoder $D$.
To avoid that the generator is incapable of fooling the discriminator, we impose a constraint that the discriminator $\mathcal{D}_s$ wins in $80\%$ of the cases \cite{gans_overwiev,sanakoyeu2018styleaware}. To achieve this, we compute a running average of the discriminator's $\mathcal{D}_s$ accuracy; if the accuracy is $<0.8$ we update the discriminator $\mathcal{D}_s$, otherwise we update the generator.

\begin{figure*}
    \begin{center}
    \includegraphics[width=\textwidth,height=\textheight,keepaspectratio]{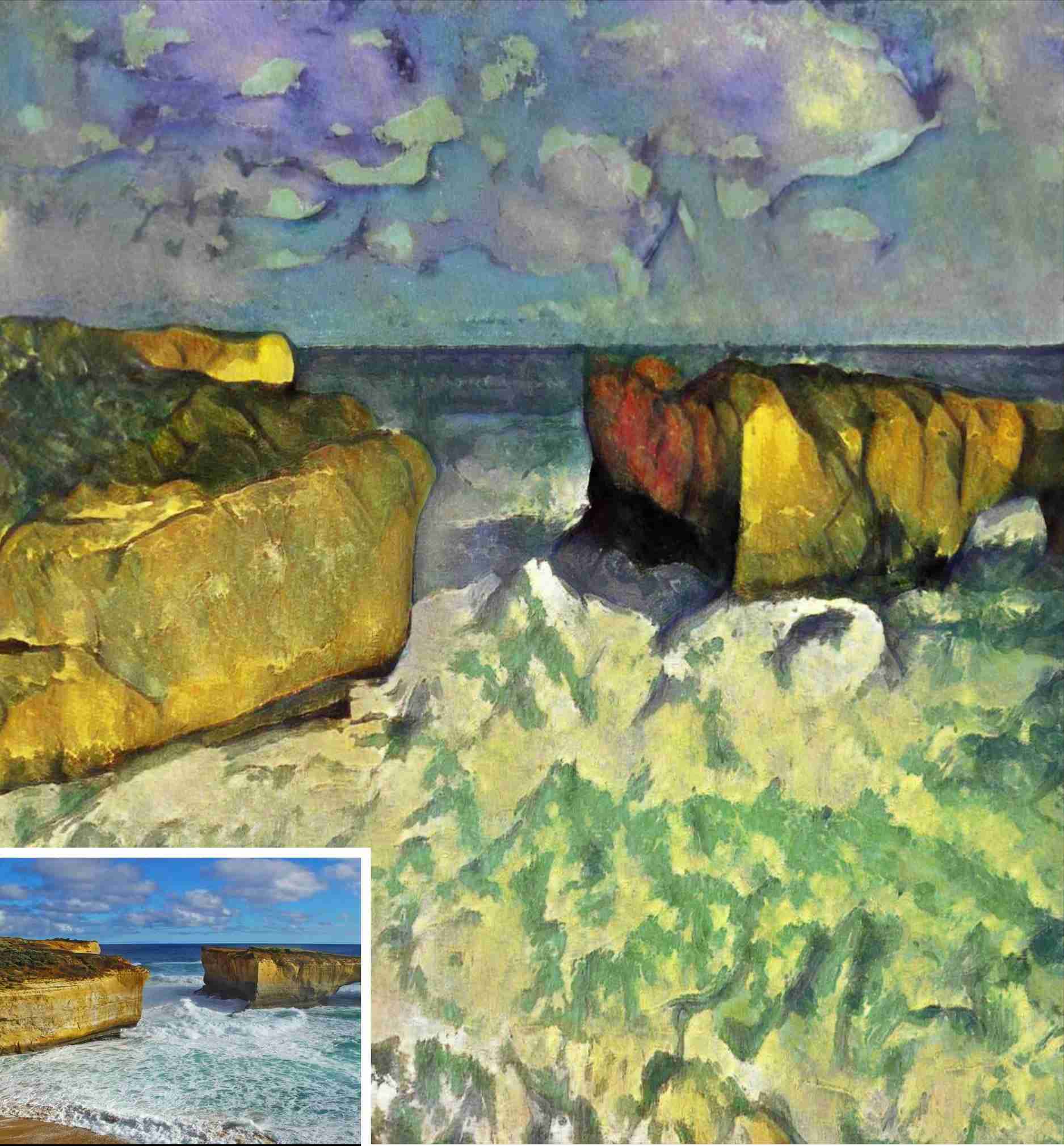}
    \caption{Image stylization for Paul Cezanne using our approach. See the \href{https://compvis.github.io/content-targeted-style-transfer/}{project page} for a hires image.}
    \label{fig:large-cezanne-1}
    \end{center}
\end{figure*}

\begin{figure*}
    \begin{center}
    \includegraphics[width=\textwidth,height=\textheight,keepaspectratio]{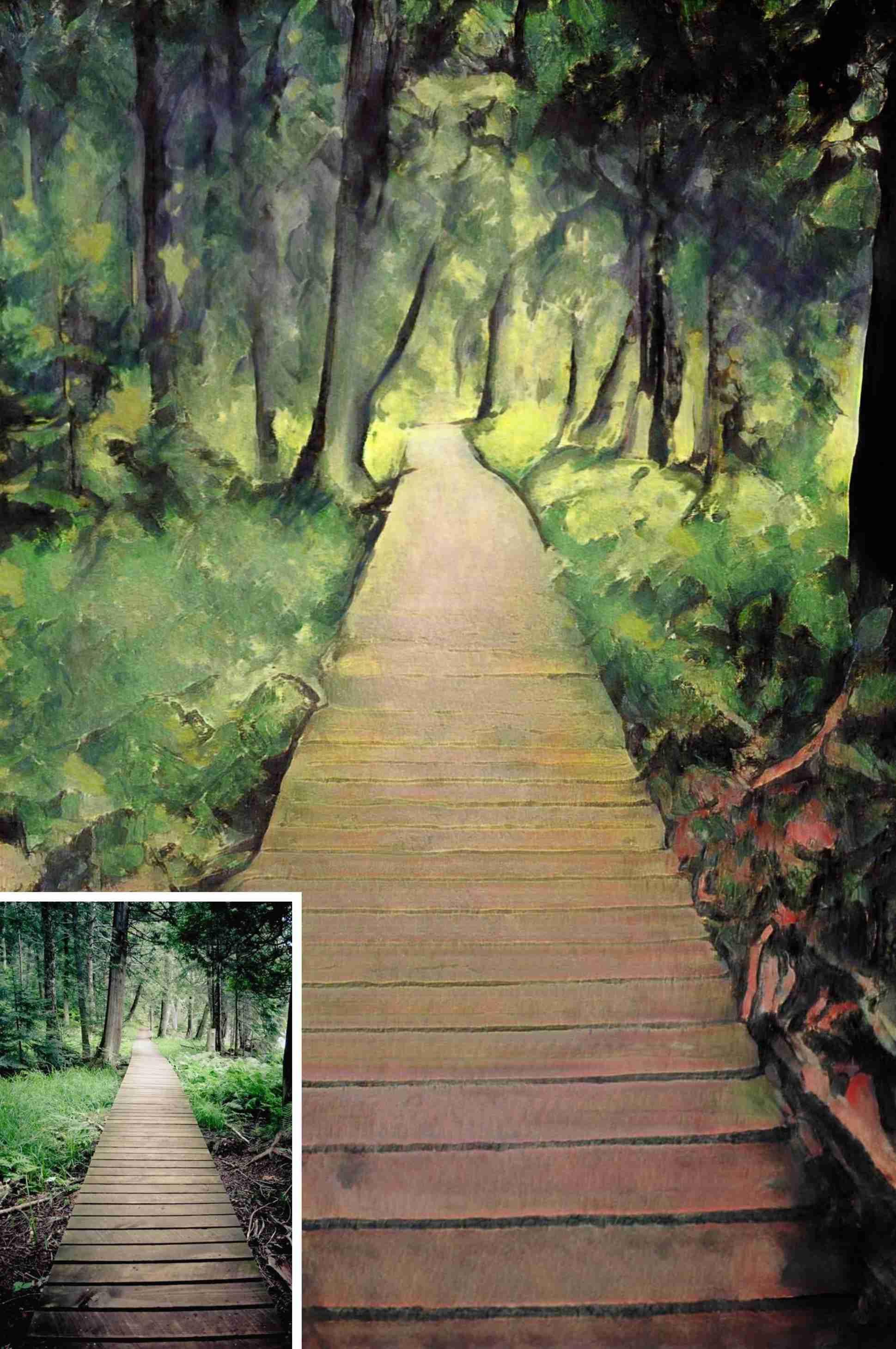}
    \caption{Image stylization for Paul Cezanne using our approach. See the \href{https://compvis.github.io/content-targeted-style-transfer/}{project page} for a hires image.}
    \label{fig:large-cezanne-2}
    \end{center}
\end{figure*}


\begin{figure*}
    \begin{center}
    \includegraphics[width=1.0\textwidth,height=1.0\textwidth]{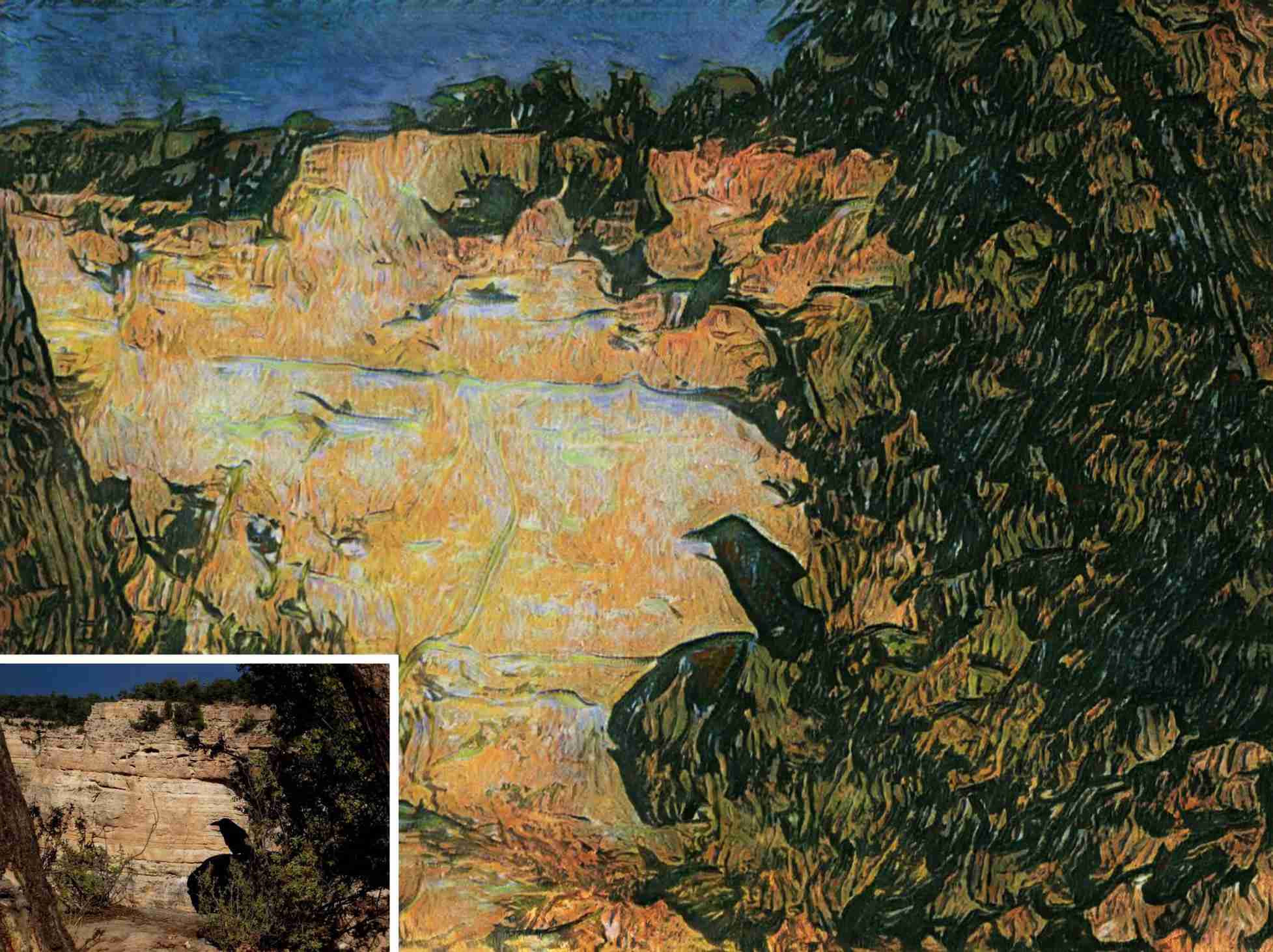}
    \caption{Image stylization for Vincent van Gogh using our approach. See the \href{https://compvis.github.io/content-targeted-style-transfer/}{project page} for a hires image.}
    \label{fig:large-gogh}
    \end{center}
\end{figure*}

\begin{figure*}
    \begin{center}
    \includegraphics[width=\textwidth,height=\textheight,keepaspectratio]{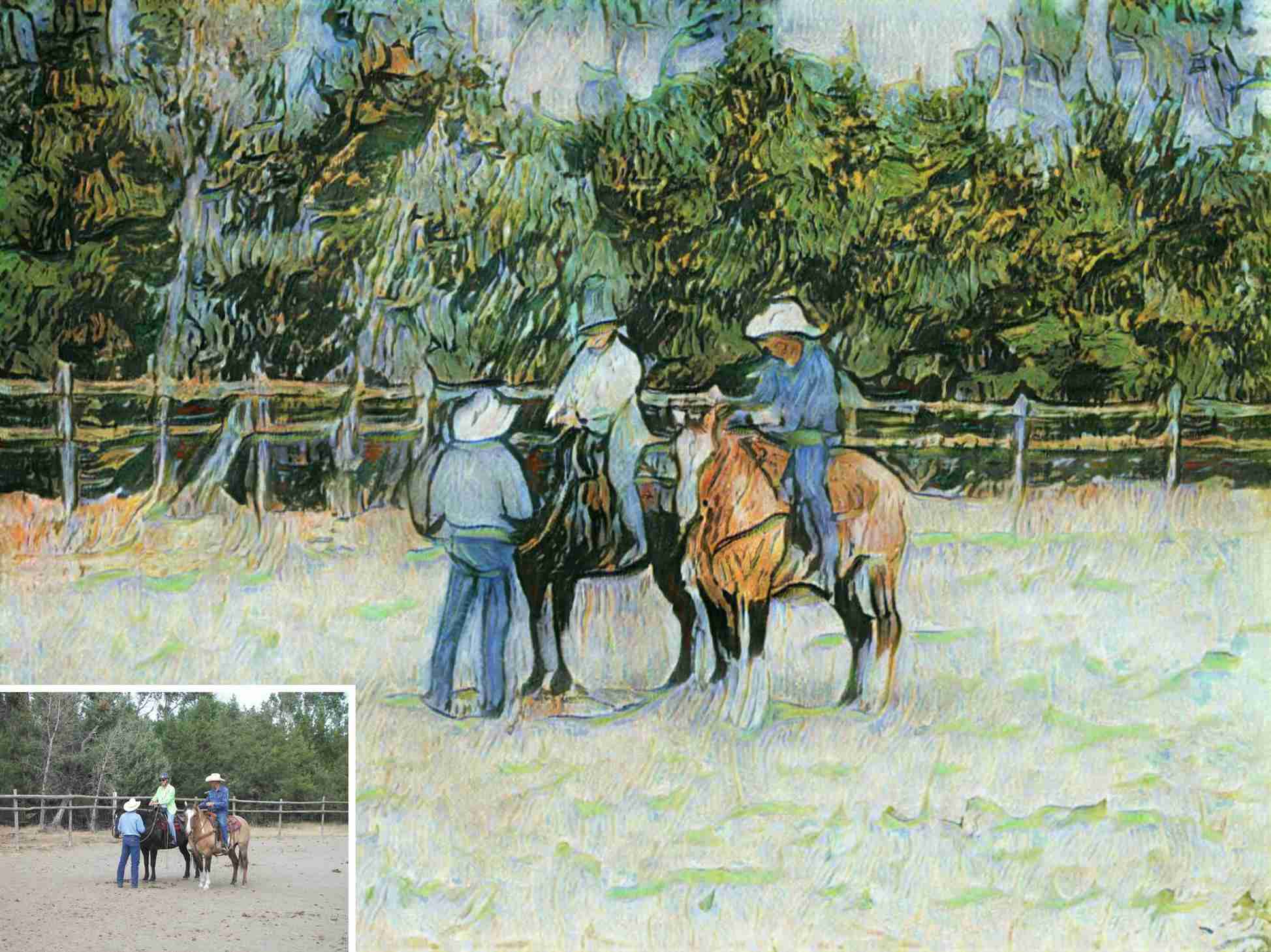}
    \caption{Image stylization for Vincent van Gogh using our approach. See the \href{https://compvis.github.io/content-targeted-style-transfer/}{project page} for a hires image.}
    \label{fig:large-vangogh-1}
    \end{center}
\end{figure*}

\begin{figure*}
    \begin{center}
    \includegraphics[width=\textwidth,height=\textheight,keepaspectratio]{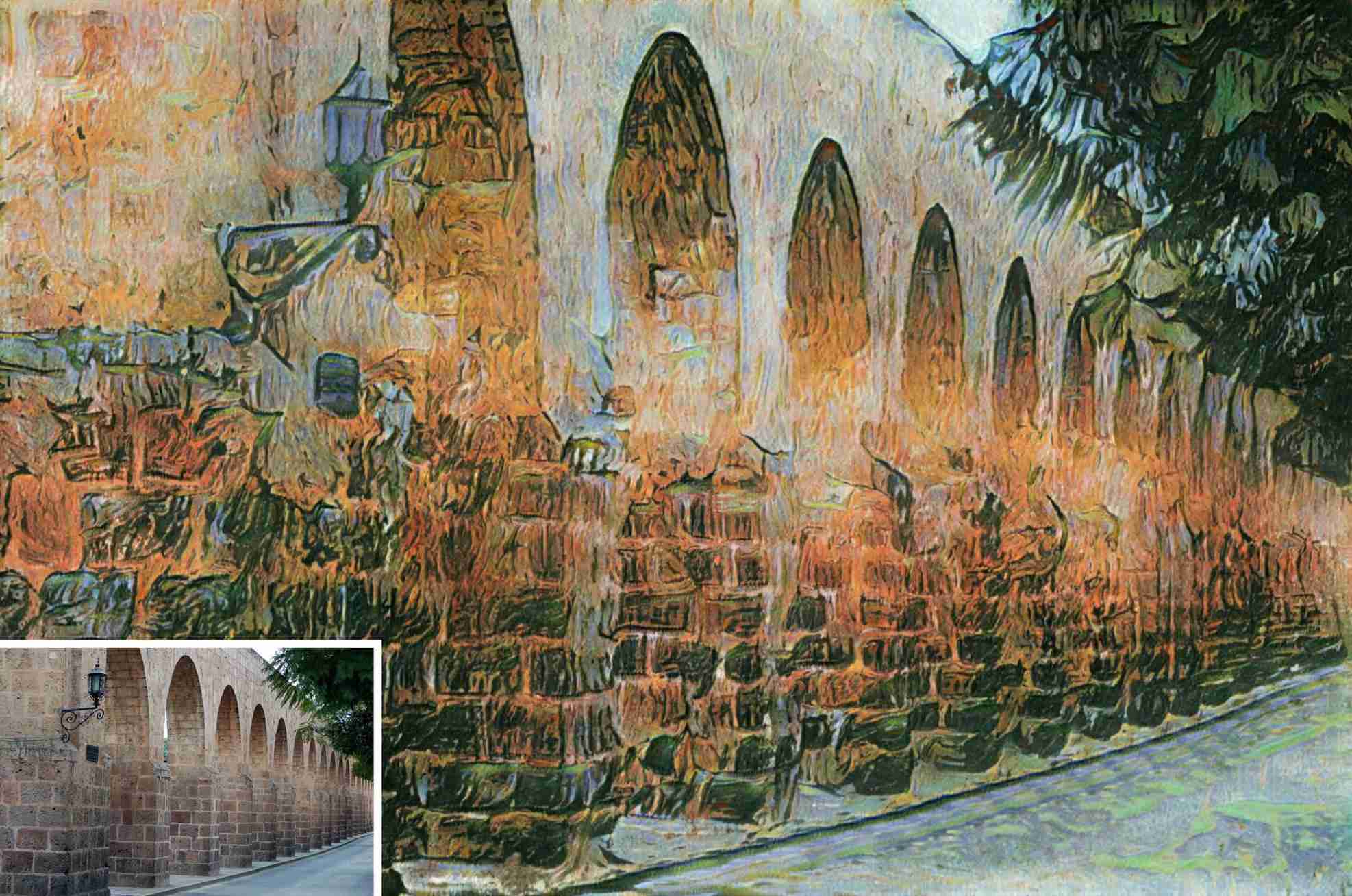}
    \caption{Image stylization for Vincent van Gogh using our approach. See the \href{https://compvis.github.io/content-targeted-style-transfer/}{project page} for a hires image.}
    \label{fig:large-vangogh}
    \end{center}
\end{figure*}

\begin{figure*}
    \begin{center}
    \includegraphics[width=\textwidth,height=\textheight,keepaspectratio]{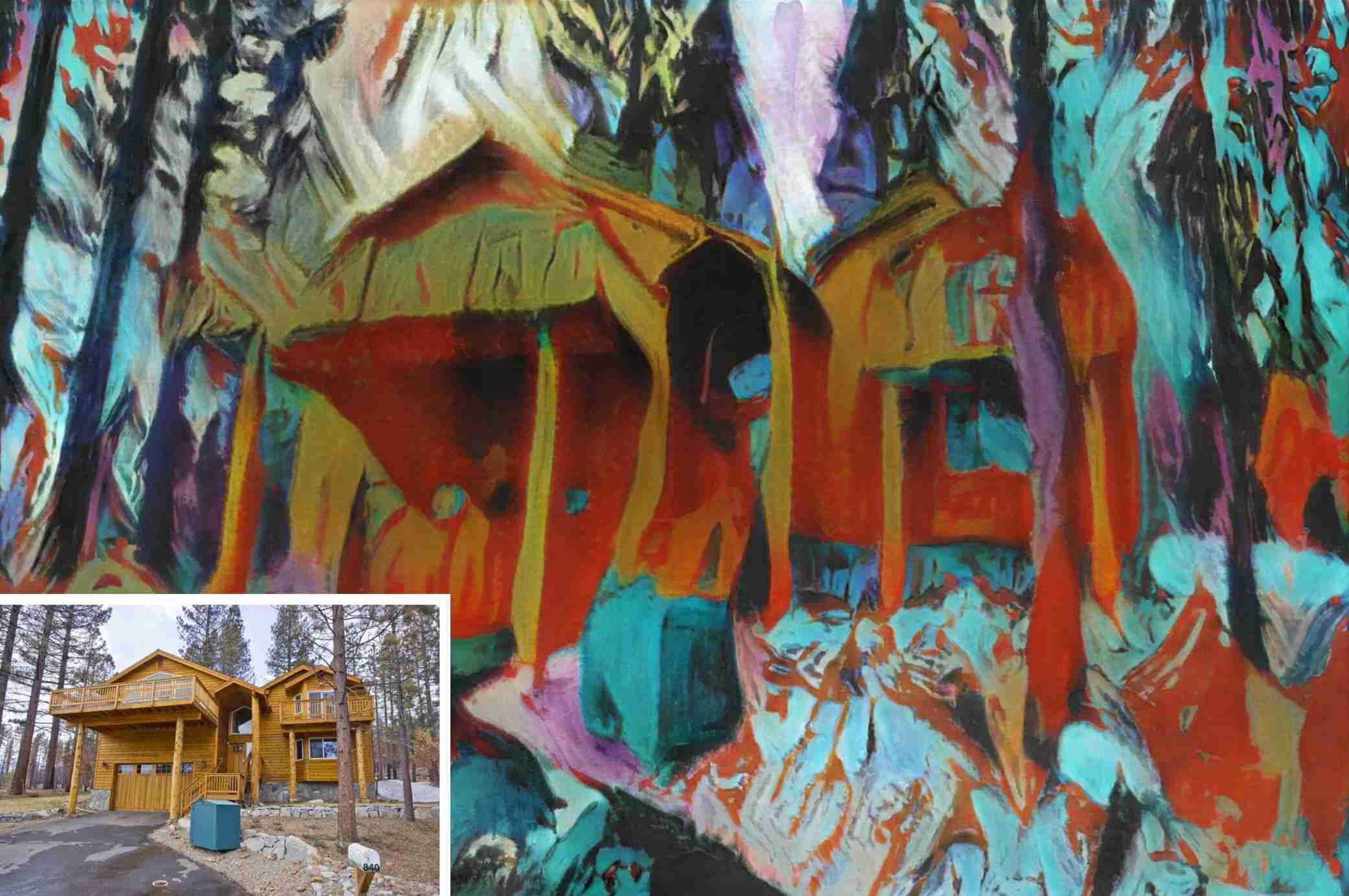}
    \caption{Image stylization for Ernst Ludwig Kirchner using our approach. See the \href{https://compvis.github.io/content-targeted-style-transfer/}{project page} for a hires image.}
    \label{fig:large-kirchner}
    \end{center}
\end{figure*}

\begin{figure*}
    \begin{center}
    \includegraphics[width=\textwidth,height=\textheight,keepaspectratio]{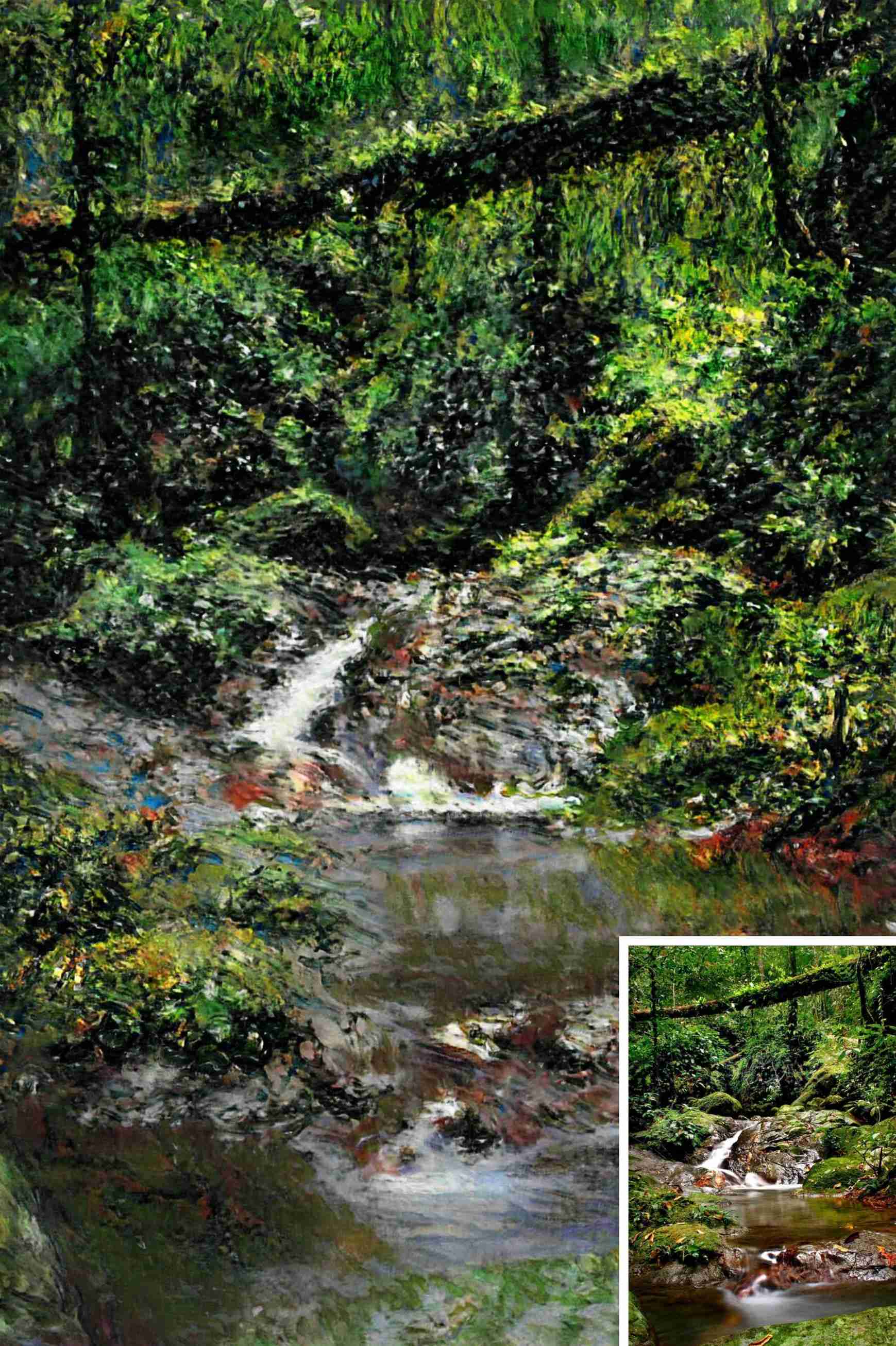}
    \caption{Image stylization for Claude Monet using our approach. See the \href{https://compvis.github.io/content-targeted-style-transfer/}{project page} for a hires image.}
    \label{fig:large-monet}
    \end{center}
\end{figure*}

\begin{figure*}
    \begin{center}
    \includegraphics[width=\textwidth,height=\textheight,keepaspectratio]{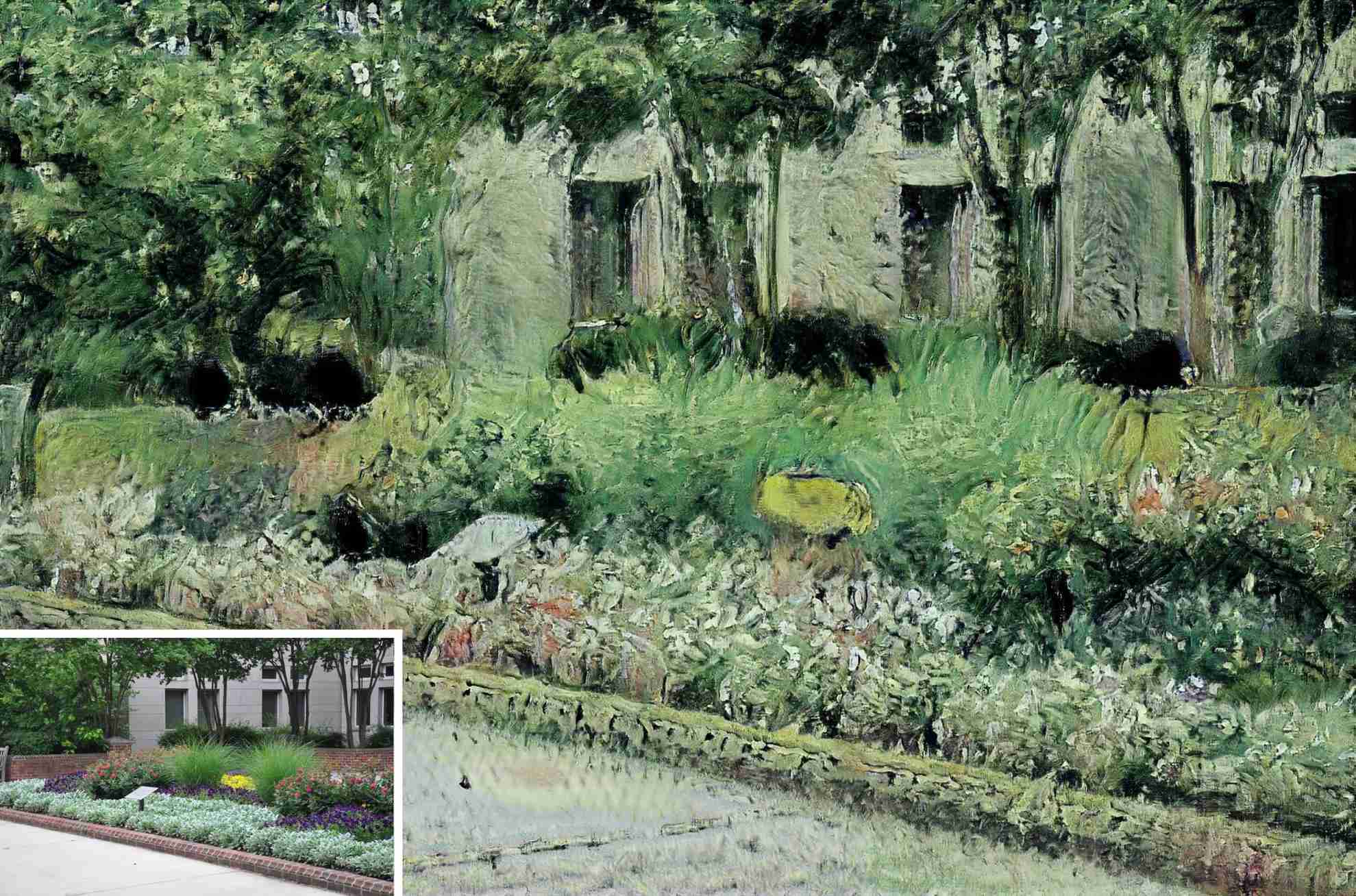}
    \caption{Image stylization for Berthe Morisot using our approach. See the \href{https://compvis.github.io/content-targeted-style-transfer/}{project page} for a hires image.}
    \label{fig:large-morisot}
    \end{center}
\end{figure*}

\begin{figure*}
    \begin{center}
    \includegraphics[width=\textwidth,height=\textheight,keepaspectratio]{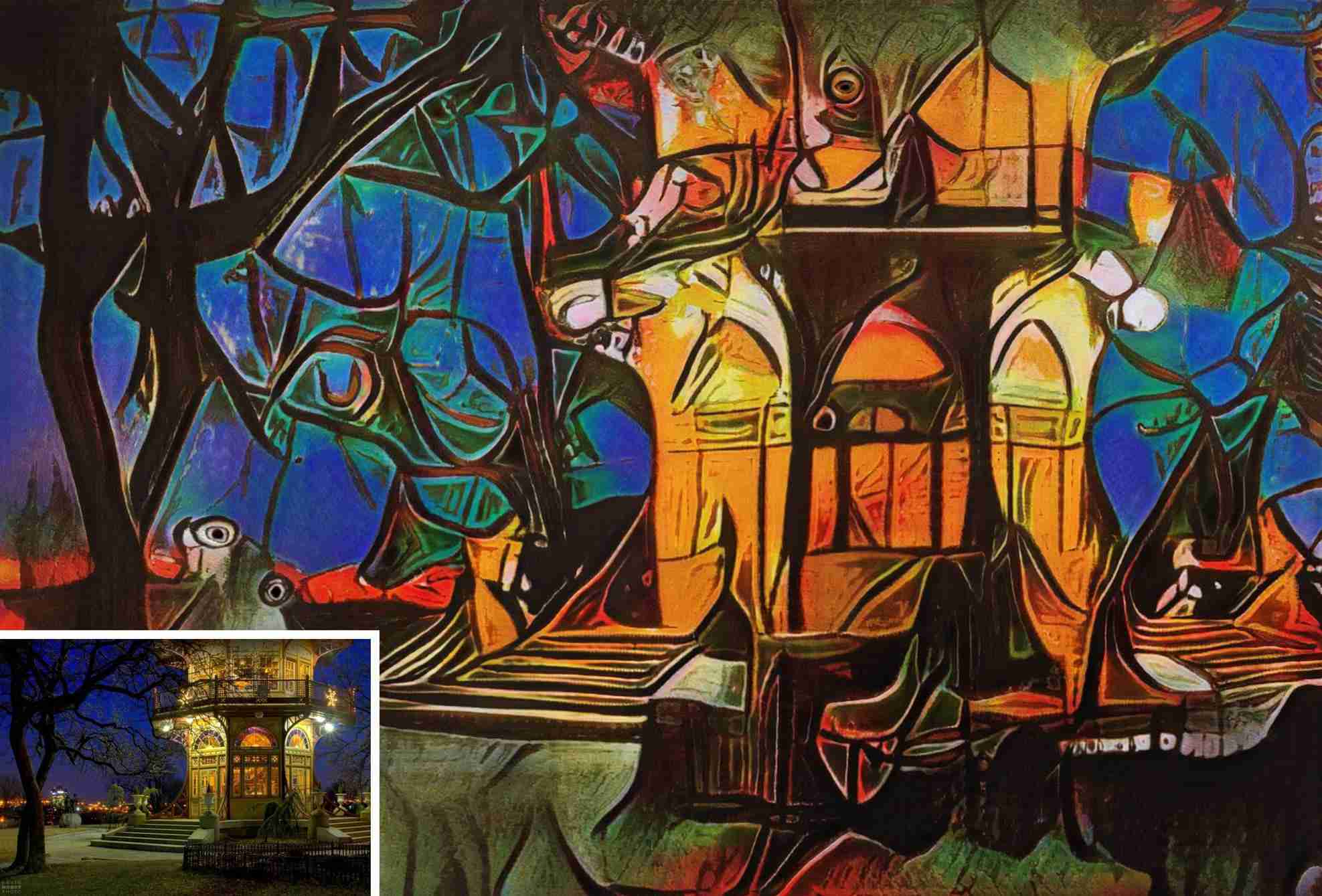}
    \caption{Image stylization for Pablo Picasso using our approach. See the \href{https://compvis.github.io/content-targeted-style-transfer/}{project page} for a hires image.}
    \label{fig:large-picasso1}
    \end{center}
\end{figure*}

\end{document}